\documentclass[lettersize,journal]{IEEEtran}
\usepackage{amsmath, amsfonts, amssymb}
\usepackage{graphicx}
\usepackage{epsfig}
\usepackage{float}
\usepackage{placeins}
\usepackage{array}
\usepackage{tabularx}
\usepackage{booktabs}
\usepackage{makecell}
\usepackage{algorithm}
\usepackage{algorithmic}
\usepackage{caption}
\usepackage{subcaption}
\usepackage{textcomp}
\usepackage{times}
\usepackage{xcolor}
\usepackage{cite}
\usepackage{url}
\usepackage[hidelinks]{hyperref}
\usepackage{verbatim}
\usepackage{balance}
\begin{document}
\title{Benchmarking Foundation Models for Zero-Shot Biometric Tasks}



\author{Redwan Sony\thanks{Corresponding authors: Redwan Sony (\texttt{sonymd@msu.edu}) and Arun Ross (\texttt{rossarun@msu.edu})}, 
        Parisa Farmanifard, 
        Hamzeh Alzwairy, 
        Nitish Shukla, 
        Arun Ross\\
        Department of Computer Science and Engineering\\
        Michigan State University, East Lansing, MI 48824}




\maketitle

\begin{abstract}
The advent of foundation models—particularly Vision-Language Models (VLMs) and Multi-modal Large Language Models (MLLMs)—has redefined the frontiers of artificial intelligence, enabling remarkable generalization across diverse tasks with minimal or no supervision. Yet, their potential in biometric recognition and analysis remains relatively underexplored. In this work, we introduce a comprehensive benchmark that evaluates the zero-shot and few-shot performance of state-of-the-art publicly available VLMs and MLLMs across six biometric tasks spanning the face and iris modalities: face verification, soft biometric attribute prediction (gender and race), iris recognition, presentation attack detection (PAD), and face manipulation detection (morphs and deepfakes).  A total of \textbf{41} VLMs were used in this evaluation. Experiments show that embeddings from these foundation models can be used for diverse biometric tasks with varying degrees of success. For example, in the case of face verification, a True Match Rate (TMR) of $96.77\%$ was obtained at a False Match Rate (FMR) of $1\%$ on the Labeled Face in the Wild (LFW) dataset, without any fine-tuning. In the case of iris recognition, the TMR@1\% FMR on the IITD-R-Full dataset was $97.55\%$ without any fine-tuning. Further, we show that applying a simple classifier head to these embeddings can help perform DeepFake detection for faces, Presentation Attack Detection (PAD) for irides, and extract soft biometric attributes like gender and ethnicity from faces with reasonably high accuracy. This work reiterates the potential of pretrained models in achieving the long-term vision of Artificial General Intelligence.

\end{abstract}


\section{Introduction}
In recent years, the field of artificial intelligence has witnessed a profound paradigm shift driven by the emergence of foundation models—large-scale models trained on broad, diverse datasets and capable of generalizing across a wide range of tasks with minimal or no task-specific supervision~\cite{bommasani2022opportunitiesrisksfoundationmodels}.  Notable examples include GPT-4~\cite{openai2024gpt4technicalreport} for language, Segment Anything Model (SAM)~\cite{kirillov2023segment_anything} for vision segmentation, MedSAM for Medical Image Segmentation~\cite{ma2024segment}  and a growing body of vision-language models (VLMs) such as CLIP~\cite{clip_radford2021learning}, ALIGN~\cite{jia2021scaling_align}, BLIP~\cite{li2022blip}, BLIP-2~\cite{li2023blip2}, Flamingo~\cite{alayrac2022flamingo},  LLaVA~\cite{llava15}, and DeepSeek-VL~\cite{lu2024deepseek_vl}.  These models are typically pretrained on billions of image-text pairs and exhibit strong zero-shot and few-shot generalization across diverse tasks without requiring task-specific supervision.

This new paradigm enables models to act as general-purpose perceptual engines. With the ability to align visual and textual representations at scale, Vision Language Models (VLMs) have shown state-of-the-art performance on benchmarks such as MS-COCO captioning~\cite{lin2014microsoft_mscoco}, VQAv2~\cite{goyal2017making}, OKVQA~\cite{marino2019okvqa}, and VizWiz~\cite{gurari2018vizwiz}. To enhance the reasoning, generalization, and scalability of foundation models—including vision-language models—a suite of advanced techniques has come to light, significantly expanding their capabilities across tasks and domains. Mixture-of-Experts (MoE) architectures selectively activate subsets of model parameters per input, offering efficient scaling with reduced computation, as demonstrated in GLaM~\cite{mixture_of_experts} and DeepSeek-VL2~\cite{wu2024deepseek_vl2}. For complex reasoning, Chain of Thought (CoT) prompting enables models to generate intermediate logical steps rather than relying on direct answer generation, greatly improving performance on multi-hop and arithmetic tasks~\cite{chain_of_thoughts}. Building on this, Graph of Thought (GoT) introduces structured reasoning paths, modeling dependencies among multiple reasoning branches~\cite{graph_of_thoughts}. Complementing these, self-consistency decoding aggregates multiple diverse reasoning trajectories to yield more reliable outputs, enhancing model robustness in ambiguous or multi-answer settings~\cite{wang2023selfconsistencyimproveschainthought}.

Incorporating external knowledge is another major direction. Retrieval Augmented Generation (RAG) combines pretrained models with retrieval modules that pull relevant documents from large corpora, improving factual accuracy and grounding~\cite{lewis2021retrievalaugmentedgenerationknowledgeintensivenlp}. Furthermore, zero-shot and few-shot learning paradigms allow foundation models to generalize to unseen tasks or domains with zero or minimal examples, often using prompts or demonstrations embedded in the input sequence~\cite{brown2020languagemodelsfewshotlearners}. This has been bolstered by in-context learning, where models are conditioned on examples at inference time without any weight updates, and instruction tuning, which fine-tunes models on a variety of tasks with natural language instructions to align outputs with user intent~\cite{sanh2022multitaskpromptedtrainingenables,chung2022scalinginstruction}. Models such as Chameleon~\cite{Chameleon2024Meta} and DeepSeek-VL2~\cite{wu2024deepseek_vl2} incorporate sophisticated architectural designs such as MoEs and unified tokenization schemes to further boost performance across tasks like image generation, captioning, grounding, and reasoning.

The recent shift from task-specific architectures toward general-purpose models is driven by both practical scalability and theoretical generalization, i.e., a model's ability to perform well on unseen data.  Traditionally, computer vision and natural language tasks required separate models, each trained from scratch or fine-tuned for a narrow objective, be it object classification, object detection, captioning, or sentiment analysis. This fragmented approach incurs high costs in terms of labeled data, domain-specific engineering, and computational resources. In contrast, foundation models consolidate learning into a single, pretrained model trained on a broad distribution of multi-modal data, enabling a unified backbone for a wide array of downstream applications~\cite{bommasani2022opportunitiesrisksfoundationmodels}. Crucially, these models encode transferable knowledge learned from massive corpora and compute-intensive training runs which sometimes reach millions of dollars, can be reused efficiently through lightweight tuning. This includes methods like instruction tuning~\cite{sanh2022multitaskpromptedtrainingenables, chung2022scalinginstruction}, which aligns model behavior with user instructions across many tasks, and reinforcement learning from human feedback (RLHF)~\cite{ouyang2022_rlhf} to fine-tune response quality based on preference data. Techniques such as Low-Rank Adaptation (LoRA)~\cite{hu2021lora} and prompt tuning~\cite{lester2021powerscaleparameterefficientprompt} further reduce the cost of model adaptation, allowing for parameter-efficient fine-tuning without retraining the full model. This paradigm democratizes access to high-performance AI: developers and researchers can now build competitive systems by adapting pretrained foundation models rather than engineering and training task-specific pipelines from scratch. Moreover, it facilitates powerful zero- and few-shot generalization abilities, enabling models to solve new problems on the fly with minimal supervision—an invaluable property in domains like biometrics, medicine, or low-resource languages, where labeled data is often scarce or sensitive.


Despite initial attempts to apply foundation models to biometric problems—such as iris segmentation using SAM (Iris-SAM)~\cite{farmanifard2024iris} and iris analysis using ChatGPT~\cite{farmanifard2024chatgpt} by Farmanifard and Ross, as well as face recognition via fine-tuned DINOv2~\cite{oquab2023dinov2} and CLIP~\cite{clip_radford2021learning} by Chettaoui et al.~\cite{chettaoui2025froundationfoundationmodelsready}—the broader potential of these models across biometric modalities remains largely underexplored. Several survey papers have explored the promise of foundation models in medicine~\cite{khan2406comprehensive}, biomedical engineering~\cite{liu2025biomedical}, medical imaging~\cite{azad2023foundational} and other fields~\cite{zhou2023comprehensive, awais2025foundation, zhang2024vision, wu2023multimodal, liang2024survey, ghosh2024exploring, chen2025unleashing}, highlighting their transformative potential across multiple domains. Narayan et al.\cite{narayan2024facexformer} proposed a unified Transformer\cite{dosovitskiy2020image_vit} model for multiple face analysis tasks.  Hatef et al.~\cite{shahreza2025foundation} conducted a comprehensive survey on foundation models in the field of biometrics. Komaty et. al.~\cite{komaty2025exploringchatgptfacepresentation} used ChatGPT for face presentation attack detection. More recently, Shekhawat et al.~\cite{shekhawat2025towards} employed ChatGPT-4.o~\cite{openai2024gpt4technicalreport} and Gemini~\cite{geminiteam2025geminifamilyhighlycapable} within a chain-of-thought framework to harness the reasoning capabilities of foundation models for differential morph attack detection. However, their approach is interactive and chat-based.  These early studies show promising adaptation in narrow settings but barely explore the full potential of vision-language models (VLMs). They offer valuable insights yet reveal a key gap—limited benchmark evaluations on real-world biometric tasks.  More importantly, there is currently no standardized benchmark or systematic evaluation framework to explore the baseline, zero-shot capabilities of multiple foundation models across a diverse range of biometric tasks. This lack of benchmarking leaves a significant gap in understanding how well VLMs can generalize to biometric-related applications, especially given the unique challenges in biometrics such as intra-class variation, cross-sensor/domain shifts, and demographic biases~\cite{nguyen2017iris, grother2019face, raji2020saving, buolamwini2018gender}.

To address this, our work takes a first step toward exploring the viability of foundation models for multiple biometric tasks by conducting a comprehensive zero-shot and few-shot evaluation across six biometric tasks in the face and iris domains. These tasks include face verification, soft biometric attribute prediction (gender and ethnicity), iris recognition, iris presentation attack detection (PAD), and face manipulation detection (morphs and deepfakes). By evaluating several publicly released state-of-the-art foundation models - in particular, VLMs without any fine-tuning on the benchmark datasets of the aforementioned tasks - we aim to establish baseline performance, uncover strengths and limitations, and shed light on the potential of foundation models in biometric tasks. 
Our findings have important implications: if VLMs demonstrate even moderate success in zero-shot biometric inference, they can serve as plug-and-play solutions—eliminating the need for task-specific retraining while leveraging the benefits of large-scale multi-modal pretraining.\footnote{However, we maintain that domain knowledge on the part of researchers is essential even if foundation models are capable of solving a broad swath of problems.} This approach opens promising avenues for building scalable, adaptable biometric systems, particularly in scenarios where labeled data is limited or privacy considerations restrict model customization.

\section{Vision-Language and Foundation Models}

\begin{figure}[!ht]
    \centering
    \includegraphics[width=0.98\linewidth]{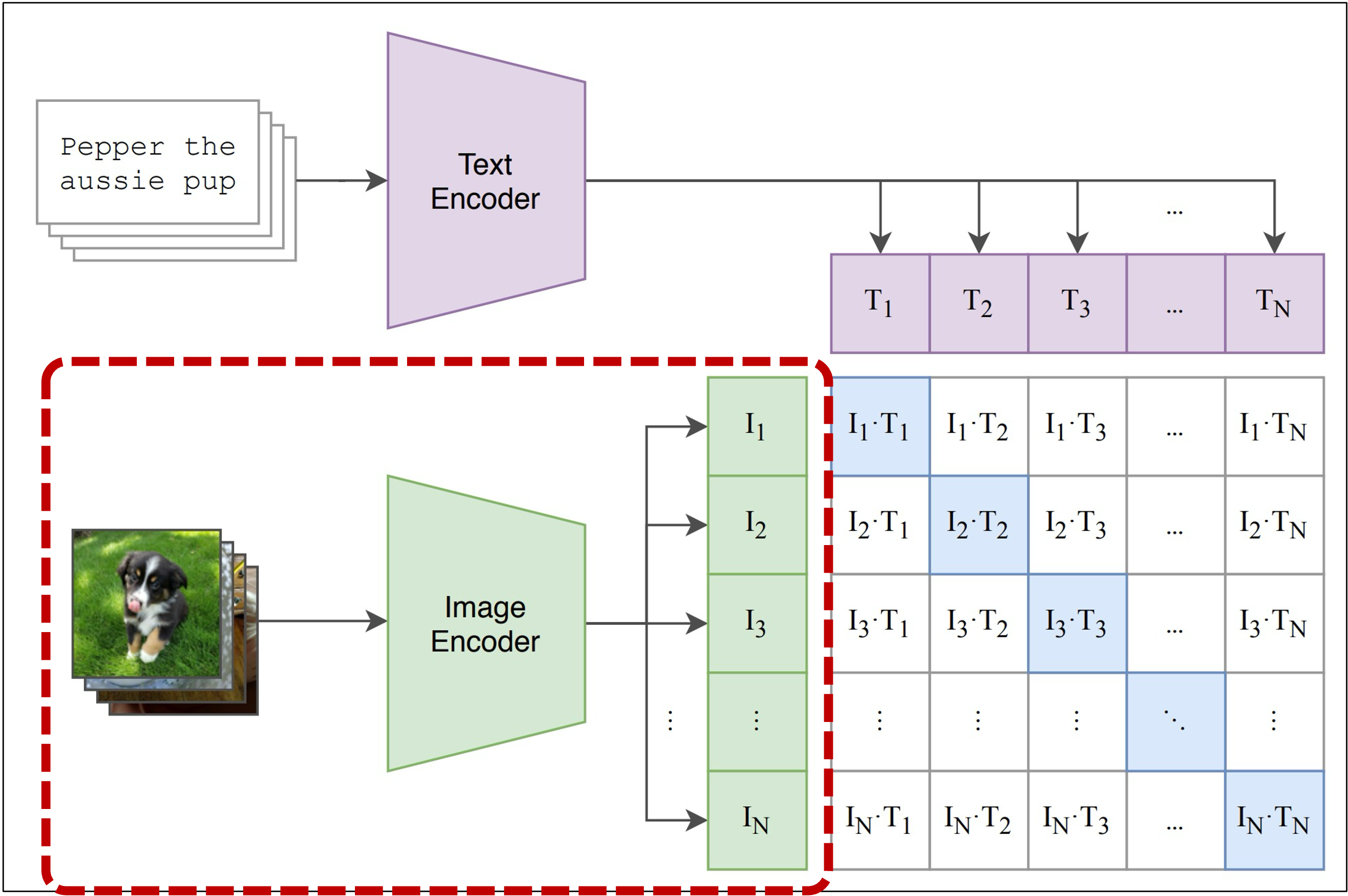}
\caption{Illustration of the CLIP model architecture, adapted from the original paper~\cite{clip_radford2021learning}. In this work, we use only the image encoder (red highlighted) to extract visual representations for biometric tasks.}
    \label{fig:clip_architecture}
\end{figure}

\subsection{CLIP (2021)}
CLIP~\cite{clip_radford2021learning}, introduced by OpenAI,\footnote{www.openai.com} is a scalable framework for learning transferable visual representations by leveraging natural language supervision. Instead of relying on fixed category labels, CLIP is trained on 400 million image–text pairs using a simple yet effective contrastive objective that aligns image and text embeddings in a shared latent space. The base CLIP models produce $768$-dimensional embeddings, while the large variants output $1,024$-dimensional features. This enables the model to understand and represent open-vocabulary visual concepts directly from text prompts. Remarkably, CLIP achieves strong zero-shot transfer across a wide range of vision tasks—matching the performance of fully supervised models like ResNet-50~\cite{he2016deep_resnet} on ImageNet~\cite{russakovsky2015imagenet}, without using any task-specific training. Its ability to generalize across domains makes it a compelling candidate for evaluating biometric recognition in a zero-shot setting. In our benchmarking, we utilize only the image encoder of the CLIP model from Figure~\ref{fig:clip_architecture} to extract image embeddings, without engaging the text encoder or the contrastive training objective.

\subsection{ALIGN (2021)}
ALIGN (A Large-scale Image and Noisy-text embedding)~\cite{jia2021scaling_align} is a vision-language model that learns aligned visual and textual representations using a dual-encoder architecture trained with contrastive learning on over 1 billion noisy image–alt-text pairs collected from the web. The model employs EfficientNet~\cite{tan19a_efficientnet} as the vision encoder and BERT~\cite{devlin2019bert} as the text encoder, jointly trained to align image and text embeddings. The ALIGN model produces $640$-dimensional feature embeddings.  Unlike prior models that rely on heavily curated datasets such as MS-COCO~\cite{lin2014microsoft_mscoco} or Conceptual Captions~\cite{sharma2018conceptual}, ALIGN demonstrates that scale can compensate for noise, achieving strong performance in zero-shot classification and setting state-of-the-art results on image-text retrieval benchmarks like Flickr30K~\cite{plummer2015flickr30k} and MS-COCO at that time. In our work, we extract the vision features from the EfficientNet backbone of the ALIGN model and benchmark them on biometric tasks following the same zero-shot evaluation protocol used for other vision-language models.

\begin{figure}[!ht]
    \centering
    \includegraphics[width=0.98\linewidth]{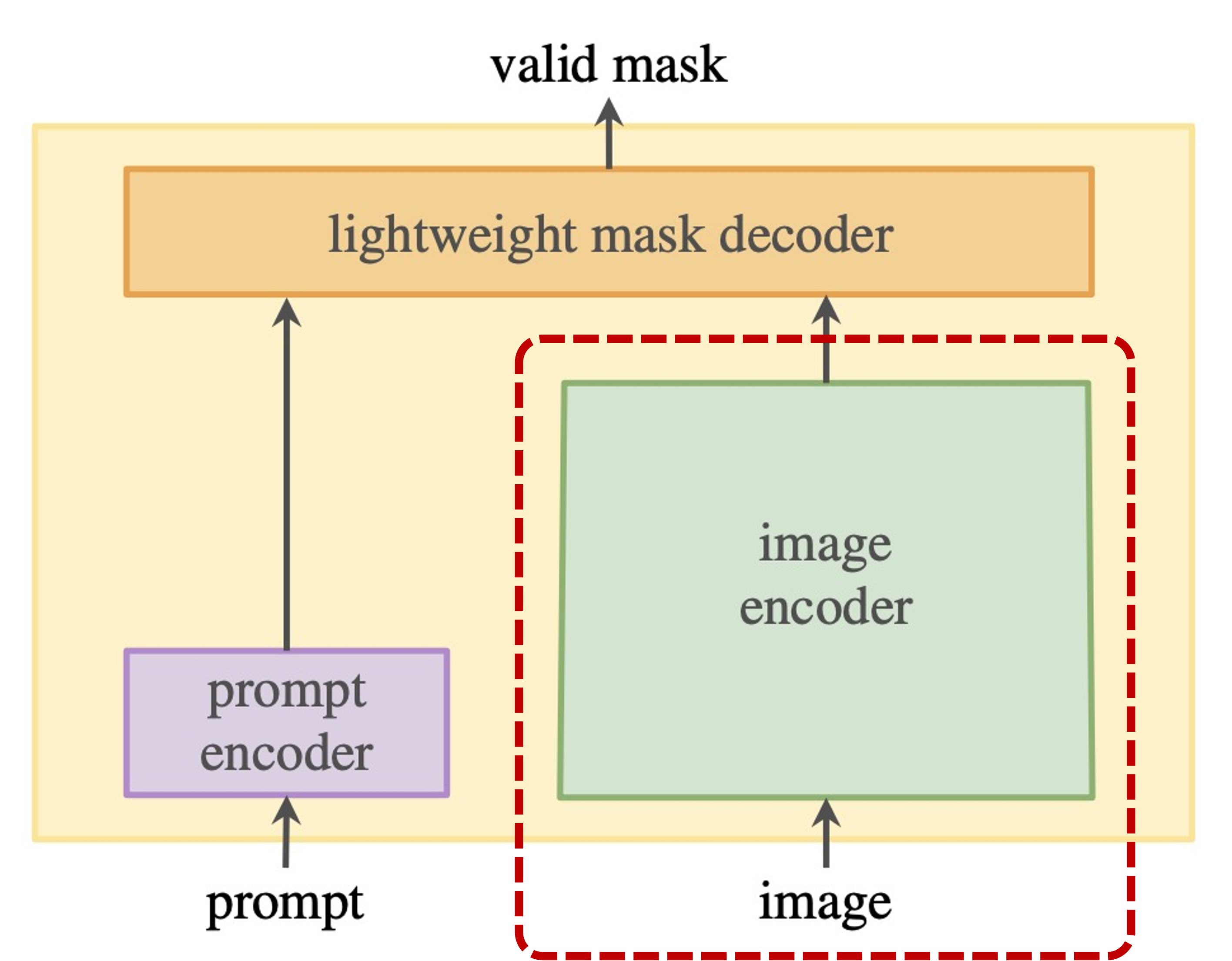}
    \caption{SAM Architecture from the original paper~\cite{kirillov2023segment_anything}. In this work, we use only the image encoder (red highlighted) to extract visual representations for biometric tasks.}
    \label{fig:sam_architecture}
\end{figure}

\subsection{OpenCLIP (2022)}
OpenCLIP~\cite{Cherti_2023_CVPR_OPENCLIP} developed by the LAION (Large-scale Artificial Intelligence Open Network)~\cite{laion_400M} community in collaboration with Hugging Face~\cite{huggingface} and other open research contributors. Unlike the original CLIP~\cite{clip_radford2021learning}, which was trained on the proprietary WIT-400M dataset, OpenCLIP models are trained on large-scale public datasets such as LAION-400M~\cite{laion_400M}, LAION-2B, and LAION-5B~\cite{LAION_5B}, making them reproducible and widely accessible. The base OpenCLIP models produce $512$-dimensional embeddings, whereas the huge and giant variants generate $1024$-dimensional feature vectors. Through systematic scaling experiments on up to 2 billion image-text pairs, OpenCLIP demonstrates that model performance follows reliable power-law scaling laws across a variety of downstream tasks, including zero-shot classification, image-text retrieval, linear probing, and fine-tuning. Importantly, the study reveals that the choice of pretraining data significantly affects model scaling behavior, even when architectures and training setups are held constant. OpenCLIP provides a strong, reproducible alternative for vision-language research and is particularly well-suited for studies on transferability, scaling, and open-world recognition.

\subsection{DINO (2021)}
DINO~\cite{caron2021dino} (short for Self-\textbf{DI}stillation with \textbf{NO} Labels) is a self-supervised learning framework that trains vision transformers (ViTs)~\cite{dosovitskiy2020image_vit} without requiring any labeled data. The method uses a teacher-student setup where both networks share the same architecture, and the teacher is updated using an exponential moving average (EMA) of the student. By matching output distributions from different augmented views of the same image, DINO enables the model to learn powerful and semantically rich representations. The DINO model produces a $768$-dimensional feature embedding. A key discovery of the paper is that self-supervised ViTs trained with DINO exhibit emergent properties—such as encoding object-level and semantic segmentation information—more explicitly than supervised ViTs or convolutional networks. The method also achieves competitive performance in linear probing and k-NN classification, reaching $80.1\%$ top-1 on ImageNet linear benchmark~\cite{russakovsky2015imagenet} using ViT-Base. In our work, we utilize the pretrained DINO vision encoder to extract image features for biometric tasks, allowing us to evaluate its zero-shot capabilities in a domain it was not originally optimized for.

\subsection{SAM (2023)}
SAM~\cite{kirillov2023segment_anything} (Segment Anything Model) is a promptable vision foundation model designed for general-purpose image segmentation. Introduced as part of the Segment Anything (SA) project, SAM is trained on SA-1B, the largest segmentation dataset to date, containing over 1 billion high-quality masks across 11 million privacy-respecting images. The core strength of SAM lies in its ability to generalize: it supports zero-shot transfer to new domains and segmentation tasks via flexible input prompts (e.g., points, boxes, or masks). Despite not being fine-tuned on specific downstream datasets, SAM often matches or surpasses fully supervised baselines in segmentation performance. Farmanifard et al.~\cite{farmanifard2024iris} fine-tuned SAM to the biometric task of iris segmentation. In our work, we use only the vision encoder of SAM as shown in Figure~\ref{fig:sam_architecture} and we use the channel-wise mean to generate $256$-dimensional feature embedding for biometric tasks. 

\subsection{DINOv2 (2023)}
DINOv2~\cite{oquab2023dinov2} is a state-of-the-art self-supervised visual representation learning framework that produces high-quality, general-purpose features without relying on labeled data. Building upon the success of DINO~\cite{caron2021dino}, DINOv2 focuses on scaling both data and model size, training Vision Transformers (ViTs)~\cite{dosovitskiy2020image_vit} with up to 1 billion parameters on a curated, diverse image dataset specifically constructed for representation learning. The DINOv2 models generate feature embeddings of varying dimensions based on their scale: \texttt{DINOv2-Small} produces $384$-dimensional embeddings, \texttt{DINOv2-Base} outputs $768$-dimensional embeddings, \texttt{DINOv2-Large} provides $1,024$-dimensional embeddings, and \texttt{DINOv2-Giant} yields $1,536$-dimensional feature vectors. Unlike prior approaches that rely on noisy web data, DINOv2 introduces a dedicated data curation pipeline to ensure semantic diversity and quality. The model is trained using self-distillation and stabilized training techniques, and its distilled variants outperform OpenCLIP~\cite{Cherti_2023_CVPR_OPENCLIP} on a wide range of image- and pixel-level benchmarks. In our work, we use the pretrained DINOv2 vision encoder to extract image features for biometric tasks.
\begin{figure}[!ht]
    \centering
    \includegraphics[width=.98\linewidth]{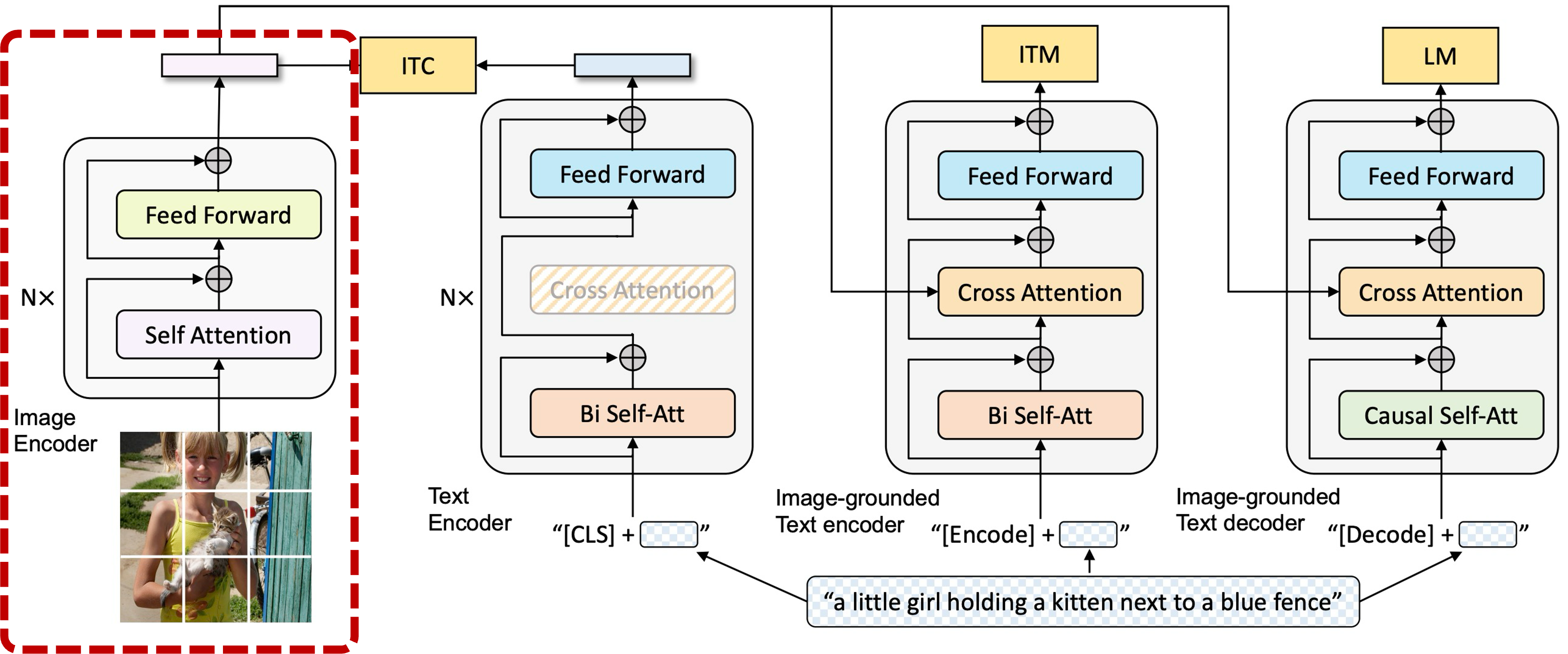}
    \caption{BLIP Architecture from~\cite{li2022blip}. In this work, we use only the image encoder to extract visual (red highlighted) representations for biometric tasks.}
    \label{fig:blip_architecture}
\end{figure}

\subsection{BLIP (2022)}
BLIP~\cite{li2022blip} introduces a flexible vision-language pretraining (VLP) framework designed to excel in both understanding-based tasks (e.g., VQA, retrieval) and generation-based tasks (e.g., image captioning). Unlike earlier VLP models that primarily rely on noisy web image-text pairs, BLIP proposes a novel bootstrapped dataset curation pipeline. This involves using a pretrained captioner to generate synthetic captions and a learned filter to remove noisy or irrelevant pairs, significantly improving the quality of supervision. The model adopts a mixture of encoder-decoder and contrastive learning architectures as shown in Figure~\ref{fig:blip_architecture}, enabling strong transferability across tasks. The base model has $768$-dimensional feature embedding where the large variant of the model uses $1,024$-dimensional embedding.  BLIP achieves state-of-the-art results across multiple benchmarks; it also demonstrates strong zero-shot generalization to video-language tasks. In our study, we use the BLIP vision encoder, highlighted in Figure~\ref{fig:blip_architecture} to evaluate its capability in biometric tasks.

\subsection{BLIP-2 (2023)}
\begin{figure}[!ht]
    \centering
    \includegraphics[width=.98\linewidth]{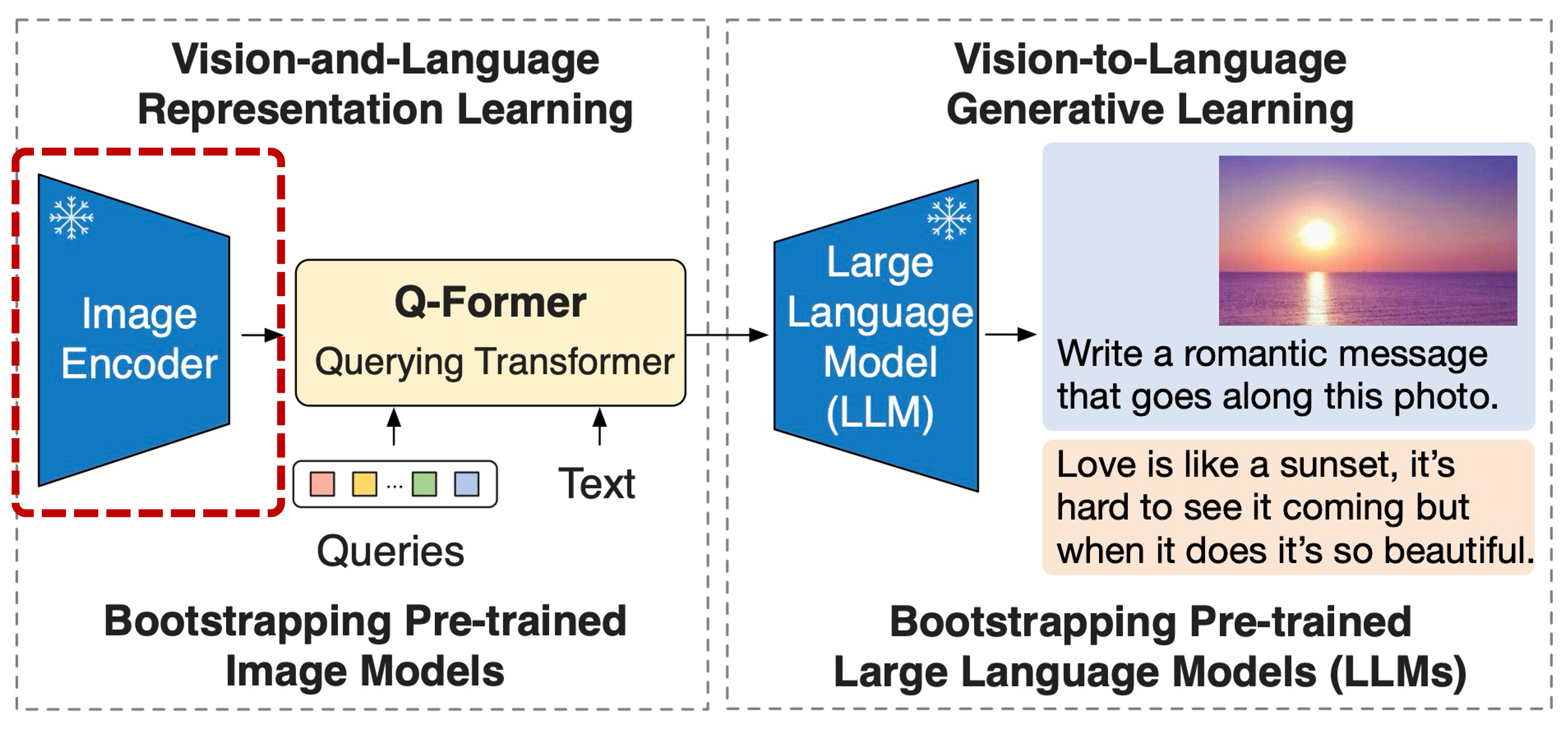}
    \caption{BLIP2 Architecture from the original paper~\cite{li2023blip2}. We use only use the image encoder to extract visual (red highlighted) representations.}
    \label{fig:blip2_architecture}
\end{figure}

BLIP-2~\cite{li2023blip2} introduces a parameter-efficient vision-language pretraining framework that dramatically reduces training cost by leveraging frozen pretrained image encoders and large language models (LLMs). Unlike traditional end-to-end training approaches, BLIP-2 bridges the vision-language modality gap using a lightweight Querying Transformer (Q-Former), which is trained in two stages: (1) vision-language representation learning with a frozen image encoder, and (2) vision-to-language generation with a frozen LLM. This modular design enables BLIP-2 to inherit strong priors from large foundation models while learning only a minimal number of task-specific parameters. It supports instruction-following image-to-text generation in a zero-shot setting. All BLIP-2 model variants produce $1,408$-dimensional feature embeddings. In this work, we evaluate the vision encoder as shown in Figure~\ref{fig:blip2_architecture} output of BLIP-2 for biometric tasks. 

\begin{figure}[!ht]
    \centering
    \includegraphics[width=.98\linewidth]{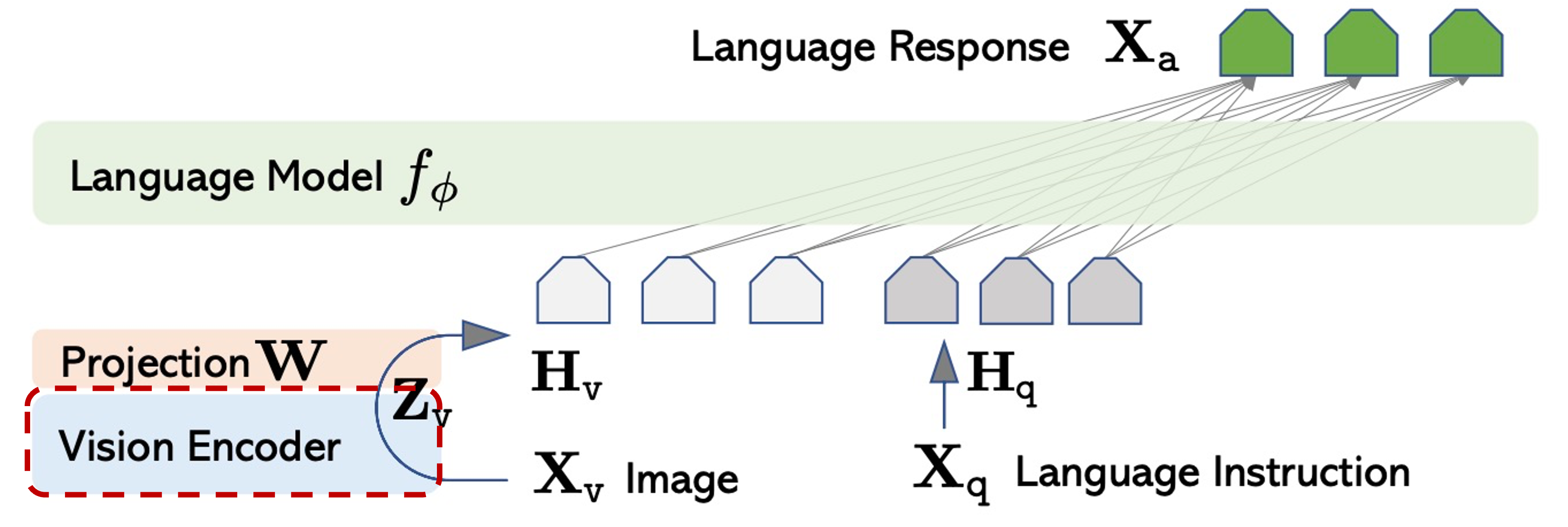}
    \caption{LLaVA Architecture from the original paper~\cite{li2023blip2}. We use only utilize the vision encoder to extract visual (red highlighted) representations.}
    \label{fig:llava_architecture}
\end{figure}

\subsection{LLaVA (2023)}
LLaVA~\cite{llava15} is a multi-modal instruction-tuned vision-language model that integrates a frozen visual encoder (e.g., CLIP ViT-L/14)~\cite{clip_radford2021learning} with an open-source large language model (Vicuna~\cite{vicuna2023} or LLaMA~\cite{touvron2023llama}) through a lightweight projection layer as shown in Figure~\ref{fig:llava_architecture}. The model is trained in two stages: (1) pre-alignment, where the image embeddings are aligned to the LLM’s language space, and (2) instruction tuning, where the entire system is trained using multi-modal instruction-response pairs. This design enables LLaVA to act as a visual assistant, capable of image-grounded dialogue, zero-shot visual question answering (VQA), and captioning through natural language prompting. Despite its relatively lightweight architecture, LLaVA achieves strong performance across multiple benchmarks, including zero-shot VQAv2~\cite{goyal2017making} and GPT-4-style multi-modal understanding. The LLaVA-1.5 model outputs $1,024$-dimensional feature embeddings, whereas the LLaVA-1.6 variant produces higher-dimensional embeddings of size $4,096$.

\subsection{Kosmos-2}
Kosmos-2~\cite{peng2023kosmos} is a multimodal large language model developed by Microsoft that processes interleaved text and image data using a unified Transformer-based architecture. Unlike traditional models that use separate modality-specific encoders, Kosmos-2 introduces visual grounding through Markdown-style annotations linking textual spans to object regions within images. Trained on the proposed GrIT~\cite{peng2023kosmos} dataset comprising grounded image-text pairs, the model enables joint reasoning over vision and language for tasks such as referring expression comprehension and grounding. In our experiments, we extract the representation corresponding to the first token (class token) from the model’s output sequence as the visual-linguistic feature embedding for downstream biometric tasks.

\subsection{Chameleon (2024)}
Chameleon~\cite{Chameleon2024Meta} is Meta’s early-fusion multimodal foundation model designed to process and generate interleaved text, image, and code sequences using a unified token space and a single Transformer backbone. Unlike late-fusion models, Chameleon tokenizes images similarly to text, enabling joint reasoning across modalities without modality-specific components. Trained on 4.4 trillion tokens, the model achieves state-of-the-art results in VQA and image captioning, outperforming Flamingo~\cite{alayrac2022flamingo} and LLaVA-1.5~\cite{llava15}, while remaining competitive on text-only benchmarks. The Chameleon model produces feature vectors of $D\times4,096$ dimensional feature vector where we use the first token (class token) as the feature vector, Here $D$ is the sequence length. 

\subsection{DeepSeek-VL (2024)}
DeepSeek-VL~\cite{lu2024deepseek_vl} is an open-source vision-language model designed for practical multimodal applications, combining strong language capabilities with detailed visual understanding. The model is trained on a diverse and realistic dataset including web screenshots, OCR, charts, and documents, with instruction tuning guided by a use-case taxonomy based on real user scenarios. It features a hybrid vision encoder optimized for high-resolution images $(1024\times1024)$ while maintaining low computational overhead, enabling effective semantic and fine-grained visual processing. Importantly, the model integrates LLM training from the beginning of pretraining, preserving robust language understanding alongside visual grounding. Both 1.3B and 7B parameter versions demonstrate competitive or state-of-the-art performance on a range of vision-language tasks while remaining efficient and responsive in chatbot settings at the time of release. In our study, we evaluate the vision encoder of DeepSeek-VL to assess its zero-shot performance in biometric recognition tasks with their $1,024$-dimensional feature embeddings.

\subsection{InternVL-3 (2024)}
InternVL3~\cite{zhu2025internvl3} is a state-of-the-art multimodal large language model (MLLM) that advances the InternVL series by adopting a native multimodal pretraining paradigm. Unlike traditional approaches that retrofit text-only LLMs with visual capabilities, InternVL3 jointly learns from both multimodal data and text corpora during a single pretraining stage, enabling better alignment and integration across modalities. The 3-billion-parameter Chameleon model generates $1,024$-dimensional feature embeddings, while the 38-billion-parameter variant produces embeddings of size $3,200$.  It introduces Variable Visual Position Encoding (V2PE) to support extended multimodal contexts and leverages post-training enhancements such as Supervised Fine-Tuning (SFT) and Mixed Preference Optimization (MPO). InternVL3 also benefits from test-time scaling and an optimized training infrastructure. The largest variant, InternVL3-78B, achieves 72.2 on the MMMU benchmark, setting a new open-source record and demonstrating competitive performance with proprietary models like GPT-4o~\cite{openai2024gpt4o}, Claude 3.5 Sonnet~\cite{anthropic2024claude35}, and Gemini 2.5 Pro~\cite{deepmind2025gemini25}. In our study, we evaluate InternVL3 for zero-shot face verification and soft-biometric attribute analysis, both through visual embedding extraction and full-chat interaction.
\begin{figure}[!ht]
    \centering
    \includegraphics[width=.99\linewidth]{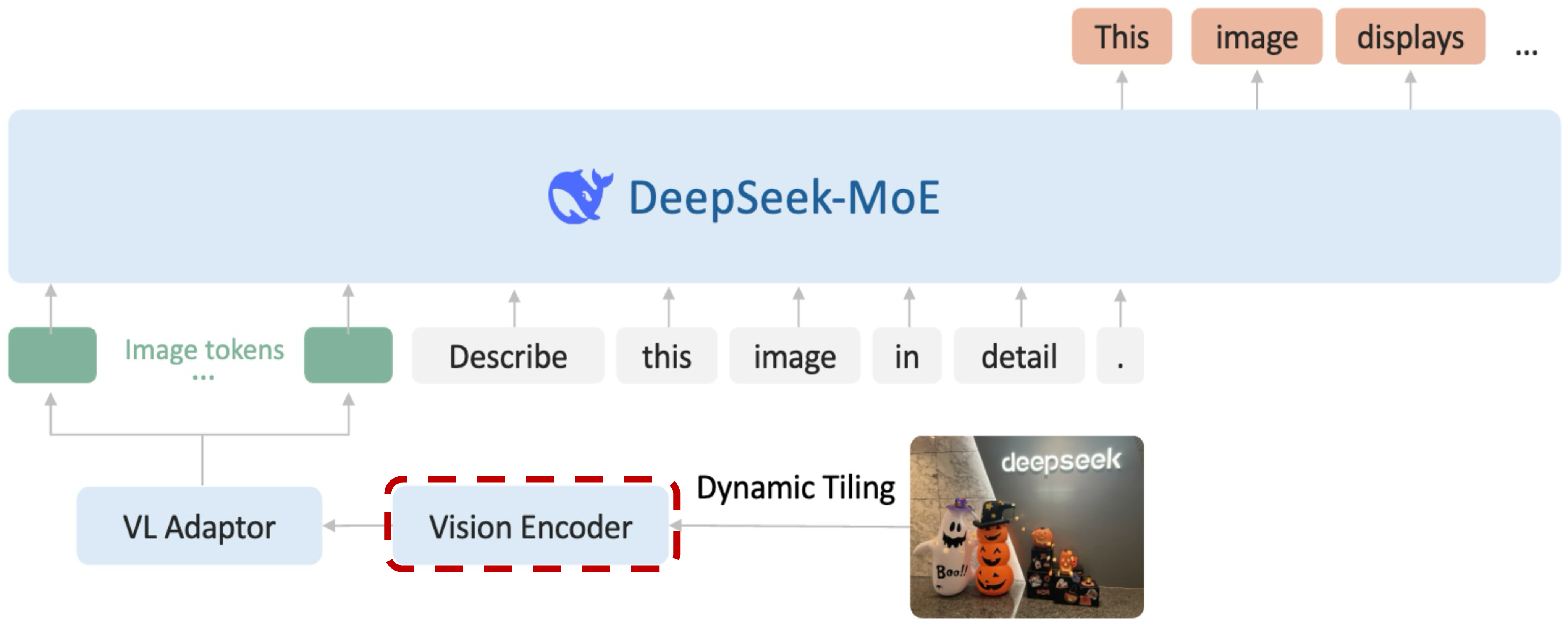}
    \caption{LLaVA Architecture from the original paper~\cite{wu2024deepseek_vl2}. We benchmark both the vision encoder for biometric tasks.}
    \label{fig:deepseekvl2_architecture}
\end{figure}

\subsection{DeepSeek-VL2 (2024)}
DeepSeek-VL2~\cite{wu2024deepseek_vl2} is an advanced Mixture-of-Experts (MoE)~\cite{mixture_of_experts}-based vision-language model that improves upon its predecessor, DeepSeek-VL~\cite{lu2024deepseek_vl}, through two major innovations: a dynamic tiling strategy for processing high-resolution images with diverse aspect ratios, and the use of DeepSeek-MoE~\cite{dai2024deepseekmoe} language models with a Multi-head Latent Attention mechanism for efficient inference and throughput. Trained on an enhanced vision-language dataset, DeepSeek-VL2 demonstrates strong performance across tasks such as visual question answering (VQA), OCR, document/chart understanding, and visual grounding. The model comes in three sizes—Tiny, Small, and Base—with 1.0B, 2.8B, and 4.5B activated parameters, respectively. Despite its compact activation footprint, DeepSeek-VL2 achieves competitive or state-of-the-art performance, highlighting its efficiency relative to other dense and MoE-based models. In our work, we evaluate both the vision encoder (as illustrated in Figure~\ref{fig:deepseekvl2_architecture}) for face embedding generation in the face recognition task, and the full DeepSeek-VL2 model as a chatbot for performing face verification and soft-biometric attribute extraction. All the variants of the DeepSeek-VL2 model produce feature vectors of $1,152$-dimension.

\section{Dataset Description}
\label{sec:dataset_description}
\subsection{Face Recognition}
We evaluate the aforementioned Vision-Language Models (VLMs) across a suite of benchmark biometric datasets to explore their zero-shot or few-shot capabilities on tasks such as face verification, iris recognition, soft-biometric classification, and presentation attack detection. For face and iris verification, we report the True Match Rate (TMR) at $1\%$ False Match Rate (FMR) as the primary evaluation metric. For soft-biometric attributes such as gender and race, we report classification accuracy, while we report both AUC and detection accuracy for our DeepFakes experiments. This section provides a brief overview of each dataset used in our evaluation.
\subsubsection{AgeDB}
\label{sec:agedb_dataset}
AgeDB~\cite{moschoglou2017agedb} is a benchmark dataset curated to evaluate the robustness of face recognition \textbf{systems under age variation,} a key challenge in real-world identity verification. It contains $12,240$ in-the-wild images of $568$ subjects, with identities spanning a wide age range from childhood to old age. The dataset provides standard face verification protocols, including AgeDB-5, AgeDB-10, and AgeDB-20, where the numbers represent the average age gap in years between pairs, and a commonly used AgeDB-30 protocol consisting of $3000$ genuine and impostor face pairs. Each protocol presents increasing difficulty levels, making AgeDB a reliable testbed for assessing age-invariant facial embeddings. In this work, we utilize the AgeDB-30 benchmark to measure the zero-shot verification performance of vision-language foundation models in handling intra-class variation due to aging.

Another version of the AgeDB dataset contains $440$ identities with annotated gender information. We use this subset for evaluating gender classification performance of the VLMs. The gender distribution is relatively balanced, with $54\%$ male and $46\%$ female subjects. Since the extracted features differ in dimensionality from the number of classes, we report 5-fold cross-validated accuracy using light-weight classifiers trained on these features. The classifiers used are K-Nearest Neighbors (KNN)~\cite{cover1967nearest}, Linear Discriminant Analysis (LDA)~\cite{fisher1936use_lda}, Logistic Regression~\cite{cox1958logiesticregression}, Ridge Regression~\cite{hoerl1970ridge}, and Support Vector Machine (SVM)~\cite{cortes1995svm}.

\subsubsection{LFW}
LFW (Labeled Faces in the Wild)~\cite{huang2008labeled} is a widely used benchmark dataset for \textbf{unconstrained face verification}, designed to evaluate recognition performance under \textbf{ real-world variations such as pose, lighting, and expression}. It contains $13,233$ face images of $5,749$ individuals, collected from the internet in uncontrolled settings. The standard verification protocol consists of $3,000$ genuine  and $3,000$ impostor pairs. 

\subsubsection{CPLFW}
CPLFW (Cross-Pose Labeled Faces in the Wild)~\cite{cplfwtech} is a challenging extension of the original LFW dataset~\cite{huang2008labeled}, specifically designed to evaluate face verification performance under pose variation. It contains the same identities as LFW but reconstructs $3,000$ genuine and $3,000$ impostor pairs with greater pose differences between the faces in each pair. By introducing larger yaw and pitch gaps, CPLFW increases intra-class variation, making it a tougher benchmark for robust and pose-invariant face recognition.

\subsubsection{CFP-FP, CFP-FF}
CFP (Celebrities in Frontal-Profile)~\cite{cfp-paper_sengupta2016frontal} is a benchmark face dataset specifically designed to evaluate \textbf{pose-invariant face verification}, particularly for \textbf{extreme pose variations}. It contains $7,000$ images of $500$ subjects, with each identity having $10$ frontal and $4$ profile images.

The dataset provides two standard protocols:
\begin{itemize}
    \item \textbf{CFP-FP} (Frontal-Profile): Evenly distributed between genuine and impostor pairs, the $7,000$ verification pairs comparing frontal vs. profile images focuses on cross-pose verification. This is a hard protocol due to the large angular difference.
    \item \textbf{CFP-FF} (Frontal-Frontal): Evenly distributed between genuine and impostor pairs, the $7,000$ verification pairs comparing frontal vs. frontal images serves as an easier intra-pose baseline.
\end{itemize}
We use both protocols to report the zero-shot face recognition performance of the vision language models.

\subsection{Gender \& Ethnicity Prediction}
For soft-biometric attribute prediction, we utilize two datasets: a gender-annotated subset of the AgeDB dataset~\cite{moschoglou2017agedb} for gender classification, and the VMER dataset~\cite{Greco_MVA2020_VMER} for ethnicity classification. The AgeDB subset contains $345$ identities with gender annotations. During cross-validation, approximately $276$ identities are used for training in each fold, while the remaining $69$ identities are reserved for testing. 

\subsubsection{VMER}
 The VGG-Face2 Mivia Ethnicity Recognition (VMER)~\cite{Greco_MVA2020_VMER} dataset is a\textbf{ large-scale ethnicity-annotated} subset of the original VGGFace2~\cite{cao2018vggface2} dataset, comprising over 3.3 million images from $9,129$ identities. Ethnicity labels were assigned via majority voting by three annotators from different ethnic backgrounds, with a fourth annotator used for tie-breaking for $500$ identities. The dataset includes four ethnicity classes and maintains modest \textbf{ gender diversity ($62\%$ male, $38\%$ female)}. The ethnicity distribution is: \textbf{Caucasian Latin} ($76.0\%$), \textbf{East Asian} ($12.4\%$), \textbf{African American} ($6.8\%$), and \textbf{Asian Indian} ($4.8\%$). To simplify benchmarking while maintaining class balance, we subsample $10$ images per identity. Since the extracted features differ in dimensionality from the number of classes, we perform 5-fold cross-validation using a light-weight classifier trained on these features. For each fold, we have $400$ identities for training and $100$ identities for testing. We report ethnicity classification accuracy in Table~\ref{table:age_gender_combined} using five methods: K-Nearest Neighbor (KNN)~\cite{cover1967nearest}, Linear Discriminant Analysis (LDA)~\cite{fisher1936use_lda}, Logistic Regression~\cite{cox1958logiesticregression}, Ridge Regression~\cite{hoerl1970ridge}, and Support Vector Machine~\cite{cortes1995svm}. Standard deviations are omitted for brevity.

\begin{figure}[!ht]
  \centering
  \begin{subfigure}[t]{0.23\columnwidth}
    \includegraphics[width=\linewidth]{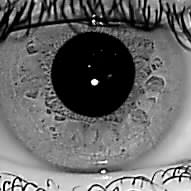}
    \caption*{\scriptsize IITD-R-Cropped\\\cite{kumar2010comparison}}
  \end{subfigure}\hfill
  \begin{subfigure}[t]{0.23\columnwidth}
    \includegraphics[width=\linewidth]{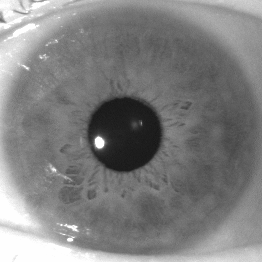}
    \caption*{\scriptsize UND-Cropped\\\cite{bowyer2008image_und_iris}}
  \end{subfigure}\hfill
  \begin{subfigure}[t]{0.23\columnwidth}
    \includegraphics[width=\linewidth]{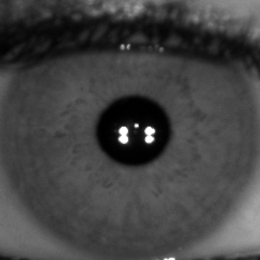}
    \caption*{\scriptsize IITD-P Normal\\Cropped~\cite{singh2018ghclnet}}
  \end{subfigure}\hfill
  \begin{subfigure}[t]{0.23\columnwidth}
    \includegraphics[width=\linewidth]{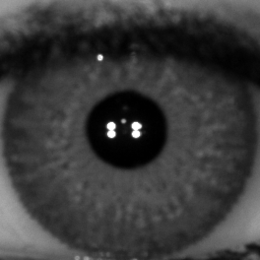}
    \caption*{\scriptsize IITD-P Patterned\\Cropped~\cite{singh2018ghclnet}}
  \end{subfigure}
  \vspace{0.5em}  

  \begin{subfigure}[t]{0.23\columnwidth}
    \includegraphics[width=\linewidth]{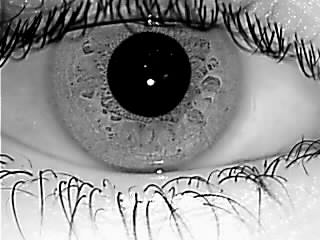}
    \caption*{\scriptsize IITD-R-Full\\\cite{kumar2010comparison}}
  \end{subfigure}\hfill
  \begin{subfigure}[t]{0.23\columnwidth}
    \includegraphics[width=\linewidth]{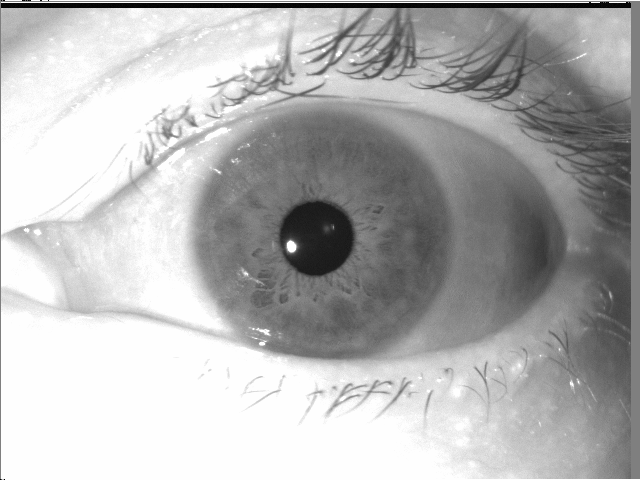}
    \caption*{\scriptsize UND-Full\\\cite{bowyer2008image_und_iris}}
  \end{subfigure}\hfill
  \begin{subfigure}[t]{0.23\columnwidth}
    \includegraphics[width=\linewidth]{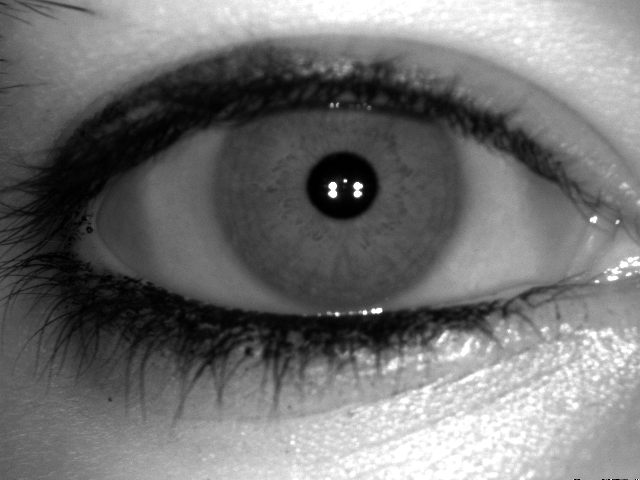}
    \caption*{\scriptsize IITD-P Normal\\Full~\cite{singh2018ghclnet}}
  \end{subfigure}\hfill
  \begin{subfigure}[t]{0.23\columnwidth}
    \includegraphics[width=\linewidth]{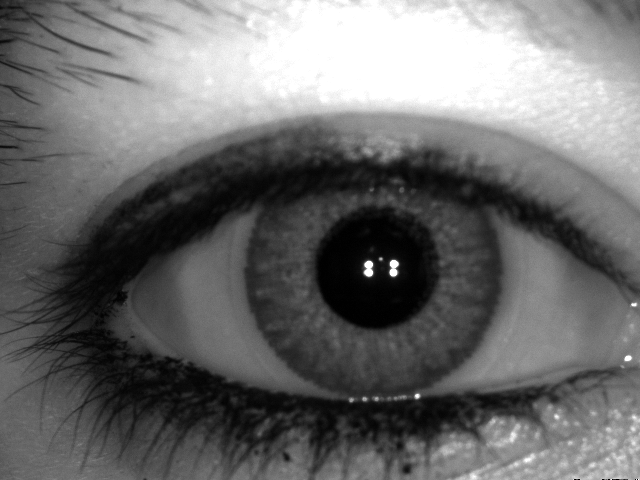}
    \caption*{\scriptsize IITD-P Patterned\\Full~\cite{singh2018ghclnet}}
  \end{subfigure}

  \caption{Example iris images from various benchmark datasets, illustrating variations in imaging conditions and sensors.}
  \label{fig:iris_datasets_row}
\end{figure}
\subsection{Iris Recognition}
\subsubsection{UND-Iris-0405}
ND-Iris-0405 dataset~\cite{bowyer2008image_und_iris}, assembled by the University of Notre Dame (UND), contains $64,980$ near-infrared (NIR) iris images collected from $356$ subjects ($712$ unique iris classes) using the LG 2200 iris imaging system, a commercial NIR sensor produced by Iridian Technologies. The sensor captures high-resolution grayscale images under controlled illumination conditions, ensuring consistent image quality suitable for iris-based analysis. For our experiments, we utilize a subset of the dataset consisting of $5,237$ images from $293$ identities, with approximately $20$ images per identity. This results in roughly $24,000$ genuine pairs and $13.7$ million impostor pairs.

\subsubsection{IIT-Delhi-Iris (IITD-R)}
IIT-Delhi Iris (IITD-R)~\cite{kumar2010comparison,iitd_iris} is a widely used benchmark dataset for evaluating iris recognition performance. This dataset was collected using the JIRIS-C1000 iris sensor under controlled acquisition conditions. It consists of $2,240$ near-infrared (NIR) iris images captured from $224$ subjects, with both left and right eye samples included. Each subject has multiple iris images, enabling the evaluation of intra-class variability. The dataset contains $1,188$ left-eye images and $1,052$ right-eye images, resulting in about $4,808$ genuine pairs and approximately $2.5$ million impostor pairs.

\subsection{Iris Presentation Attack}
\subsubsection{IIT-Delhi-Iris (IITD-P)}
The IIT-Delhi Contact Lens Iris Database~\cite{singh2018ghclnet} consists of $6,570$ iris images acquired from $101$ subjects using two commercial iris sensors: the Cogent CIS 202 dual iris sensor and the Vista FA2E single iris sensor. The dataset includes iris images captured under various conditions, with and without contact lenses. In our presentation attack detection (PAD) experiments, we selected a subset of $1,557$ iris images from the full dataset. This subset was limited to two classes based on the presence or absence of patterned contact lenses: \textbf{Patterned} and \textbf{Normal}. It includes samples from all $101$ identities, with $784$ iris images labeled as patterned and $773$ as normal,  with a total of $779$ left-eye images and $778$ right-eye images. This binary subset was used to evaluate the performance of vision-language models in distinguishing between authentic irides and those wearing patterned contact lenses. For the zero-shot experiments, we used the entire subset without any training. For classifier-based evaluation, we followed an identity-aware split, using 80\% of the identities for training and the remaining 20\% for testing.

\subsection{DeepFake Detection}
\subsubsection{FaceForensics++}
FaceForensics++ \cite{roessler2019faceforensicspp} is a DeepFake dataset composed of 1,000 original video sequences that have been manipulated using four different deepfake methods: \textbf{DeepFakes} \cite{DeepFakesGitHub}, \textbf{Face2Face }\cite{thies2020face2facerealtimefacecapture}, \textbf{FaceSwap} \cite{KowalskiFaceSwap}, and \textbf{NeuralTextures} \cite{thies2019deferredneuralrenderingimage}. The videos were collected from YouTube and primarily feature frontal facial views.

\begin{figure}[ht]
    \centering
    \includegraphics[width=0.98\linewidth]{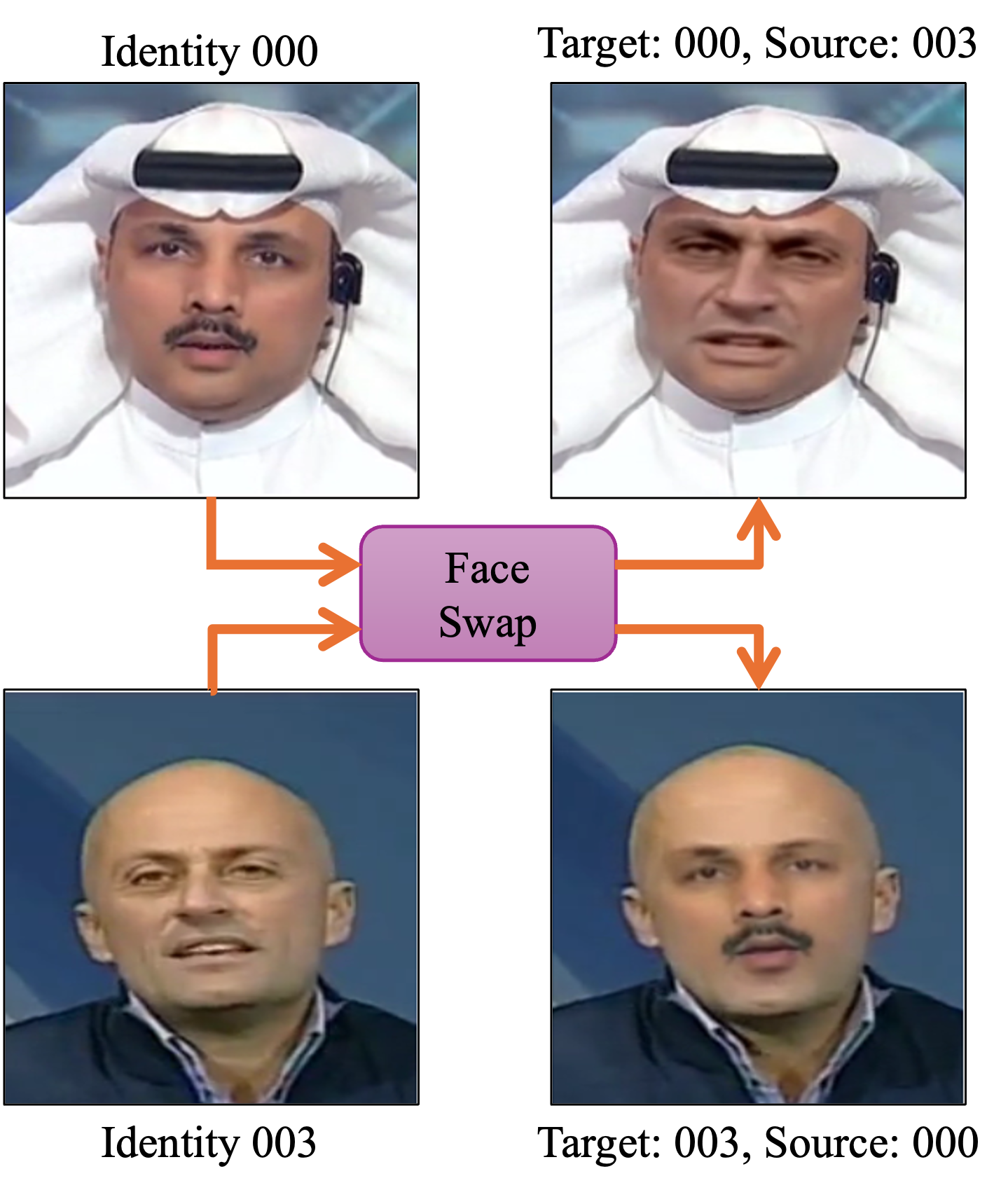}
    \caption{An example of the identity pairing in FaceForensics++ \cite{roessler2019faceforensicspp} using the DeepFakes~\cite{DeepFakesGitHub} method.}
    \label{fig:deepfake_vs_real}
\end{figure}

\begin{figure*}[!ht]
    \centering
    \includegraphics[width=\linewidth]{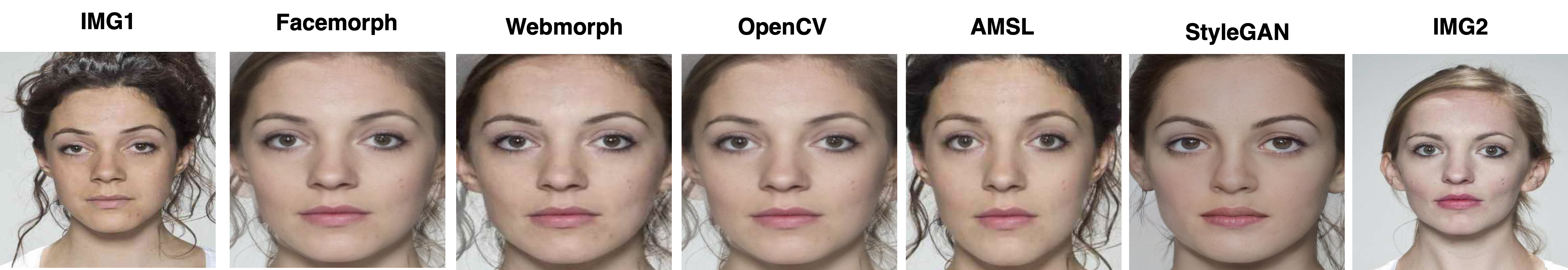}
    \caption{Morphs created from face images IMG1 and IMG2. Webmorph, OpenCV and Facemorph and AMSL are landmark-based morphs whereas StyleGAN and Mordiff are deep-learning based morphs.  }
    \label{fig:morph}
\end{figure*}

For each manipulation method, \textbf{$1,000$} fake videos are generated by pairing different identities—each pair consisting of a target (the person whose face is manipulated) and a source (the person providing the manipulated portion). For instance, video \textbf{000.mp4} corresponding to identity \textit{000}  might be paired with video \textbf{003.mp4} corresponding to identity \textit{003}, where identity \textit{000} serves as the target and identity \textit{003} as the source, and vice-versa. An example is shown in Figure \ref{fig:deepfake_vs_real}.
To ensure no identity overlap between training and testing, we split the dataset by identity pairs—so the model is never exposed to test-set identities during training. We construct the training set by extracting frames from all training videos, resulting in \textbf{$10,000$} samples. Similarly, we extract \textbf{2,000} frames from the test videos to create the testing set; the video frames were cropped to a $256\times256$ frame centered around the face and we use the \textbf{$10,000$} examples for training and the \textbf{$2,000$} for testing.
Note that in our experiments, we focus only on the three primary manipulation methods: DeepFakes~\cite{DeepFakesGitHub}, Face2Face~\cite{thies2020face2facerealtimefacecapture}, and FaceSwap~\cite{KowalskiFaceSwap}. We exclude NeuralTextures~\cite{thies2019deferredneuralrenderingimage}, as it primarily performs lip-sync-based manipulations that are difficult to detect without analyzing multiple consecutive frames.

\subsection{Morph Attack Detection}
A facial morph is a composite image created from two face images corresponding to two different identities such that the morph matches with other images of both constituent identities. We conduct MAD (Morph Attack Detection) experiments on six commonly used face morph datasets, namely, AMSL \cite{ref64}, FRLL-Morphs \cite{ref65}, and MorDiff \cite{ref9}. The FRLL-Morphs dataset consists of morphs generated using four different techniques: OpenCV \cite{ref97}, StyleGAN \cite{ref69}, WebMorph \cite{ref70}, and FaceMorph \cite{ref68}. To train the classifier, we employ the strategy discussed in \cite{dcgan,shukla2025metric}, i.e., we create training morphs from synthetic faces in the SMDD dataset \cite{ref20} using the OpenCV/dlib morphing algorithm \cite{ref97}. We create $5,000$ training morphs from $15,000$ synthetic face images in the SMDD dataset. In all test datasets, the non-morphed images are sourced from the FRLL dataset, which comprises $102$ identities, each with two frontal images—one smiling and one neutral—totaling $204$ face images. The number of morphed images in each dataset is as follows: AMSL – $2,175$; FaceMorpher – $1,222$; StyleGAN – $1,222$; OpenCV – $1,221$; WebMorph – $1,221$; and MorDiff – $1,000$. The chosen test datasets encompass both traditional landmark-based approaches and modern generative methods (See Figure \ref{fig:morph}).

\begin{table*}[!ht]
\centering
\caption{TMR (\%) at 1\% FMR across multiple datasets for different vision-language models. The best accuracy for each benchmark is bolded.}
\resizebox{1.00\textwidth}{!}{  
\begin{tabular}{l|r|rrrr|r|r|r|r}
\Xhline{1.5px}
\multicolumn{10}{c}{\textbf{Face Recognition Results}}\\
\Xhline{1.5px}
\textbf{Model} & \textbf{Param} & \multicolumn{4}{c|}{\textbf{AgeDB}~\cite{moschoglou2017agedb}} & \textbf{CPLFW} & \textbf{LFW} & \textbf{CFP\_FP} & \textbf{CFP\_FF}\\
\cmidrule(lr){3-6}
&\textbf{Billions}& \textbf{5 Year}  & \textbf{10 Year} & \textbf{20 Year}& \textbf{30 Year}  & ~\cite{cplfwtech} & ~\cite{huang2008labeled} & ~\cite{cfp-paper_sengupta2016frontal} & ~\cite{cfp-paper_sengupta2016frontal} \\
\Xhline{2pt} 
InternVL3-1B~\cite{zhu2025internvl3}&0.93& 14.10 & 7.53 & 4.73 & 1.40 & 7.57 & 37.17 & 18.77 & 33.23 \\
InternVL3-38B~\cite{zhu2025internvl3} &38.39& 11.27 & 7.57 & 3.60 & 1.93 & 7.17 & 47.50 & 21.86 & 31.46 \\\hline

BLIP-Base~\cite{li2022blip} &0.25&14.03 & 8.10 & 5.30 & 1.70 & 7.60 & 33.87 & 29.29 & 38.94 \\
BLIP-Large~\cite{li2022blip}&0.47&14.37 & 8.00 & 4.43 & 2.23 & 6.03 & 46.40 & 32.43 & 45.17 \\\hline

BLIP2-t5-xl~\cite{li2023blip2} &3.94& 63.47 & 48.03 & 29.23 & 16.43 & 43.00 & 96.20 & 87.37 & 95.54 \\
BLIP2-t5-xl-coco~\cite{li2023blip2} &3.94& 51.73 & 36.13 & 24.83 & 9.30 & 26.93 & 90.40 & 76.66 & 89.80 \\
BLIP2-t5-xxl~\cite{li2023blip2} &12.23& 63.33 & 47.93 & 29.23 & 16.70 & 43.03 & 96.27 & 87.40 & \textbf{95.60} \\
BLIP2-O-2.7b~\cite{li2023blip2} &3.74& 63.50 & 48.17 & 29.27 & 16.40 & 43.00 & 96.30 & 87.37 & 95.57 \\
BLIP2-O-6.7b~\cite{li2023blip2} &7.75& 63.33 & 48.20 & 29.17 & 16.40 & 43.33 & 96.27 & 87.46 & 95.54 \\\hline

DeepSeekVL-1.3B~\cite{lu2024deepseek_vl} &1.98& 21.07 & 16.30 & 7.77 & 3.97 & 1.57 & 50.07 & 41.37 & 55.03 \\
DeepSeekVL-7B~\cite{lu2024deepseek_vl} &7.34& 21.00 & 15.83 & 7.37 & 3.47 & 8.93 & 50.00 & 5.91 & 5.17 \\\hline

DeepSeek-vl2~\cite{dai2024deepseekmoe}&27.48& 24.93 & 16.23 & 6.73 & 3.87 & 13.40 & 67.67 & 41.51 & 58.83 \\
DeepSeek-vl2-S~\cite{dai2024deepseekmoe} &16.15& 25.00 & 15.37 & 6.43 & 3.13 & 13.67 & 63.37 & 42.26 & 58.80 \\
DeepSeek-vl2-T~\cite{dai2024deepseekmoe} &3.37& 20.23 & 13.47 & 5.57 & 2.43 & 11.97 & 49.93 & 31.34 & 51.89 \\\hline

DINO-ViTb16~\cite{caron2021dino} &0.09& 8.33 & 4.27 & 3.87 & 3.03 & 3.30 & 15.37 & 9.83 & 18.20 \\
DINOv2-Base~\cite{oquab2023dinov2} &0.02& 10.80 & 7.53 & 4.77 & 2.20 & 5.73 & 29.30 & 13.74 & 16.29 \\
DINOv2-Giant~\cite{oquab2023dinov2} &1.10& 14.03 & 7.73 & 4.47 & 2.10 & 6.37 & 27.70 & 10.34 & 11.51 \\
DINOv2-Large~\cite{oquab2023dinov2} &0.30& 13.17 & 6.27 & 3.37 & 1.40 & 5.10 & 31.30 & 11.20 & 12.91 \\
DINOv2-Small~\cite{oquab2023dinov2} &0.02& 7.87 & 6.70 & 3.83 & 1.70 & 5.63 & 24.73 & 12.31 & 17.89 \\\hline

SAM-ViT-Base~\cite{kirillov2023segment_anything} &0.90& 3.83 & 2.77 & 1.50 & 1.60 & 1.93 & 5.80 & 3.69 & 6.66 \\
SAM-ViT-Huge~\cite{kirillov2023segment_anything} &0.64& 3.23 & 3.50 & 2.00 & 1.10 & 1.87 & 4.93 & 2.34 & 8.00 \\
SAM-ViT-Large~\cite{kirillov2023segment_anything} &0.31& 4.03 & 3.27 & 2.57 & 1.70 & 1.23 & 5.33 & 3.00 & 7.86 \\\hline

ViT-B-16~\cite{dosovitskiy2020image_vit} &0.09& 10.33 & 5.90 & 4.93 & 2.47 & 4.57 & 28.73 & 19.80 & 19.40 \\
ViT-B-16-384~\cite{dosovitskiy2020image_vit} &0.09& 12.10 & 7.00 & 4.20 & 2.33 & 5.27 & 30.60 & 18.74 & 18.23 \\
ViT-B-32-384~\cite{dosovitskiy2020image_vit} &0.09& 9.90 & 6.60 & 4.80 & 1.87 & 5.33 & 27.87 & 18.83 & 20.20 \\
ViT-H-14-224-21k~\cite{dosovitskiy2020image_vit} &0.64& 8.23 & 5.87 & 4.70 & 1.60 & 3.30 & 18.20 & 14.17 & 23.66 \\
ViT-Hybrid-384~\cite{dosovitskiy2020image_vit} &0.91& 12.77 & 8.93 & 3.80 & 2.00 & 8.13 & 41.07 & 17.17 & 24.00 \\
ViT-L-16-224~\cite{dosovitskiy2020image_vit} &0.31& 9.37 & 7.43 & 3.57 & 1.53 & 4.57 & 26.77 & 18.37 & 26.57 \\
ViT-L-16-384~\cite{dosovitskiy2020image_vit} &0.31& 10.73 & 6.80 & 2.97 & 2.37 & 4.57 & 30.37 & 19.80 & 24.89 \\
ViT-L-32-384~\cite{dosovitskiy2020image_vit} &0.31& 10.07 & 6.07 & 4.17 & 1.83 & 5.13 & 27.17 & 16.94 & 18.94 \\\hline

OpenCLIP-B-32~\cite{Cherti_2023_CVPR_OPENCLIP}&0.15& 38.70 & 27.03 & 16.37 & 6.30 & 20.30 & 87.00 & 57.29 & 76.06 \\
OpenCLIP-H-14~\cite{Cherti_2023_CVPR_OPENCLIP} &0.99& 58.23 & 47.70 & 32.47 & 20.83 & 35.67 & \textbf{96.77} & 64.71 & 83.03 \\
OpenCLIP-G-14~\cite{Cherti_2023_CVPR_OPENCLIP} &1.37& 54.40 & 42.13 & 31.30 & 16.40 & 33.00 & 94.73 & 68.17 & 82.57 \\\hline


ALIGN~\cite{jia2021scaling_align} &0.17& 19.53 & 15.53 & 8.53 & 3.83 & 5.93 & 50.67 & 27.97 & 43.57 \\\hline

LLaVA-1.5-7B~\cite{llava15}&7.06& \textbf{64.47} & 53.83 & 34.13 & \textbf{21.23} & 33.67 & 91.73 & 86.20 & 92.63 \\
LLaVA-1.6-M-7b~\cite{liu2024llavanext} &7.57& 11.23 & 6.70 & 3.40 & 1.00 & 4.03 & 35.37 & 13.00 & 33.09 \\\hline

CLIP-B-16~\cite{clip_radford2021learning} &0.15& 51.00 & 37.10 & 23.00 & 10.83 & 26.87 & 89.23 & 76.74 & 88.86 \\
CLIP-B-32~\cite{clip_radford2021learning} &0.15& 39.43 & 28.77 & 12.23 & 7.70 & 19.57 & 84.07 & 66.20 & 79.14 \\
CLIP-L-32~\cite{clip_radford2021learning} &0.43& 63.17 & 49.73 & 34.23 & 21.00 & 32.77 & 91.73 & \textbf{87.63} & 93.11 \\
CLIP-L-14-336~\cite{clip_radford2021learning} &0.43& 64.40 & \textbf{54.03} & \textbf{34.87} & 21.03 & 33.67 & 91.73 & 86.20 & 92.54 \\\Xhline{2pt} 
\end{tabular}}
\label{tab:face_verification_tmr@fmr}
\end{table*}

\section{Methodology and Experimental Setup}
\subsection{Face Recognition (Verification)}

For the face recognition task, we adopt a verification-based evaluation protocol. We utilize the vision encoder part of different vision-language foundation models to extract  the feature embeddings from face images. Verification is performed by computing the cosine similarity between embeddings of image pairs, distinguishing between genuine (same identity) and impostor (different identity) pairs.

For the Labeled Faces in the Wild (LFW) Celeb dataset~\cite{huang2008labeled}, we follow the standard protocol comprising $3,000$ genuine pairs and $3,000$ impostor pairs. For the AgeDB dataset~\cite{moschoglou2017agedb}, we employ the predefined age-gap protocol, which includes verification pairs with age differences of 5, $10$, $20$, and $30$ years. While the original protocol consists of $10$ folds with $600$ pairs each (balanced between genuine and impostor pairs), we aggregate all folds and report the consolidated results.
The Celebrity Frontal-Profile (CFP) dataset~\cite{cfp-paper_sengupta2016frontal} includes two evaluation protocols: frontal-to-frontal (CFP-FF) and frontal-to-profile (CFP-FP). Each protocol is divided into $10$ folds, with $350$ genuine and $350$ impostor pairs per fold. Finally, the Cross-Pose LFW (CPLFW) dataset~\cite{cplfwtech} presents a more challenging verification task with $3,000$ genuine and $3,000$ impostor pairs, specifically curated to introduce significant pose variations between image pairs. For all the aforementioned face recognition datasets, we report performance in terms of the True Match Rate at $1\%$ False Match Rate \textbf{(TMR@1\%FMR)} in Table~\ref{tab:face_verification_tmr@fmr}.

\subsection{Soft Biometric Feature Extraction}
We assess the capability of foundation vision-language models in capturing soft biometric attributes, such as gender and ethnicity. To evaluate gender classification, we extract face embeddings from the AgeDB dataset~\cite{moschoglou2017agedb} and train light-weight classifiers i.e., Logistic Regression~\cite{cox1958logiesticregression}, Ridge Regression~\cite{hoerl1970ridge}, K-Nearest Neighbour (KNN)~\cite{cover1967nearest}, Linear Discriminant Analysis (LDA)~\cite{fisher1936use_lda} and Support Vector Machine (SVM)~\cite{cortes1995svm} on top of the extracted features. The evaluation is conducted using 5-fold cross-validation to ensure robustness. For ethnicity classification, we employ the VMER dataset~\cite{Greco_MVA2020_VMER} and follow a similar procedure. We extract features using the vision encoder of each model and train a shallow classifier to predict ethnicity labels, again validated using 5-fold cross-validation. To maintain clarity and conciseness, we report only the average classification accuracy across folds and omit the standard deviations. The results are included in Table~\ref{table:age_gender_combined}.

\subsection{Morph Attack Detection}
We benchmark several vision-language models (VLMs) across six datasets comprising both traditional landmark-based and deep learning-based morphs. From these datasets, we extract embeddings for $60,000$ bona fide face images and $30,000$ morph images using the models listed in Table \ref{tab:eer_bpcer}. For each model, we train a decision tree classifier and evaluate its performance on all six datasets. The evaluation is conducted using two metrics: the Equal Error Rate (EER) and the Bona Fide Presentation Classification Error Rate (BPCER) at a 10\% threshold of the Attack Presentation Classification Error Rate (APCER).

\section{Results}
\subsection{Face Recognition}

We evaluate the aforementioned  Vision-Language Models (VLMs) on the task of face recognition, specifically in a verification setting. For each of the benchmark face recognition datasets described in Section~\ref{sec:dataset_description}, we extract face embeddings using the vision module of each VLM and compute cosine similarity scores between pairs, without any additional training. We report the resulting zero-shot face verification performance as the True Match Rate (TMR) at 1\% False Match Rate (FMR) for each dataset in Table~\ref{tab:face_verification_tmr@fmr}.

\begin{figure*}[!ht]
    \centering
    \begin{subfigure}[t]{0.32\textwidth}
        \centering
        \includegraphics[width=\linewidth]{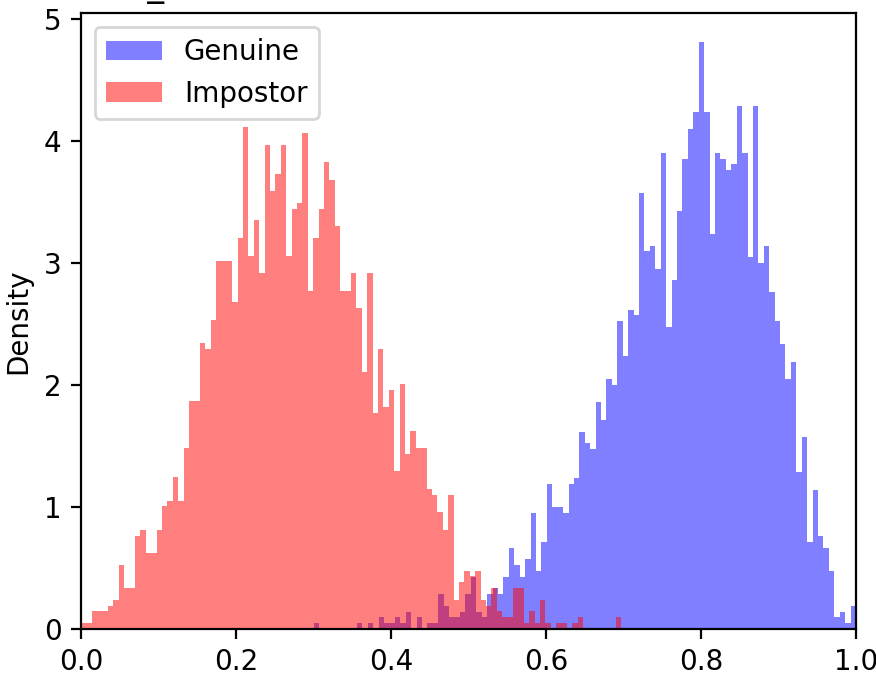}
        \caption*{OpenCLIP~\cite{Cherti_2023_CVPR_OPENCLIP}, TMR: $96.77\%$}
    \end{subfigure}
    \hfill
    \begin{subfigure}[t]{0.32\textwidth}
        \centering
        \includegraphics[width=\linewidth]{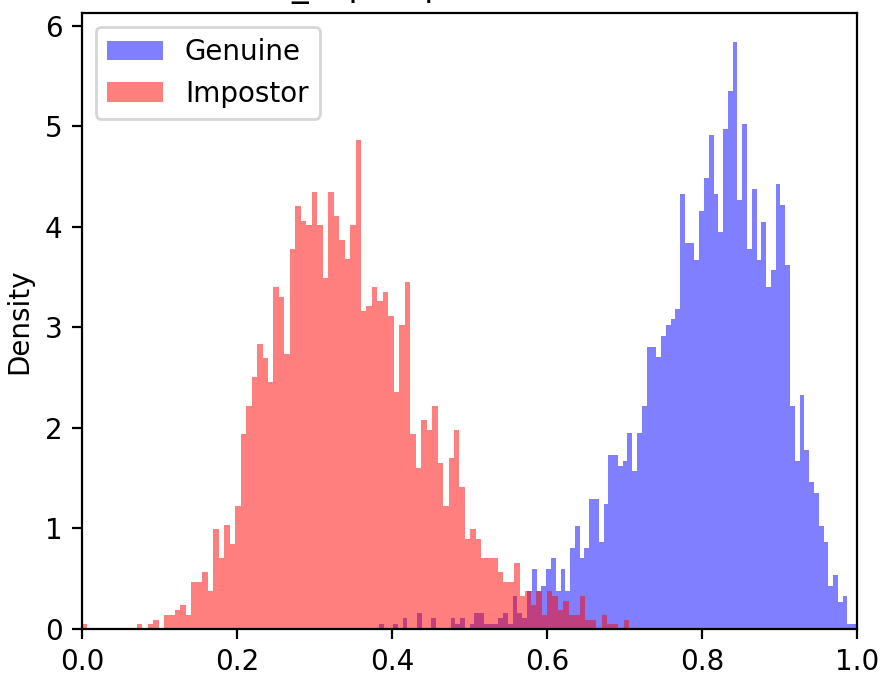}
        \caption*{BLIPv2~\cite{li2023blip2}, TMR: $96.30\%$}
    \end{subfigure}
    \hfill
    \begin{subfigure}[t]{0.32\textwidth}
        \centering
        \includegraphics[width=\linewidth]{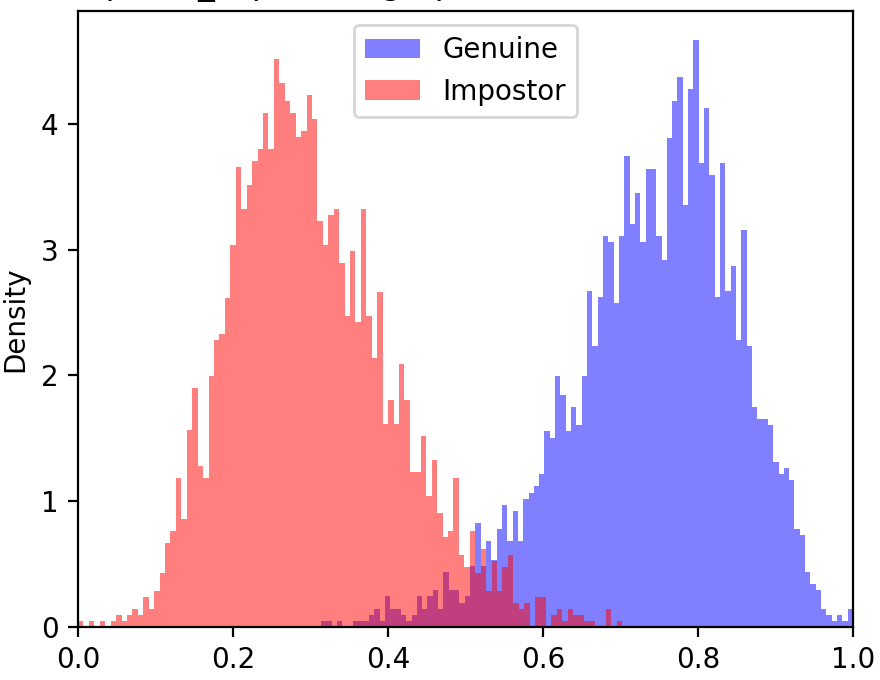}
        \caption*{CLIP-H-14~\cite{clip_radford2021learning}, TMR: $91.73\%$}
    \end{subfigure}

    \vspace{1em} 

    \begin{subfigure}[t]{0.32\textwidth}
        \centering
        \includegraphics[width=\linewidth]{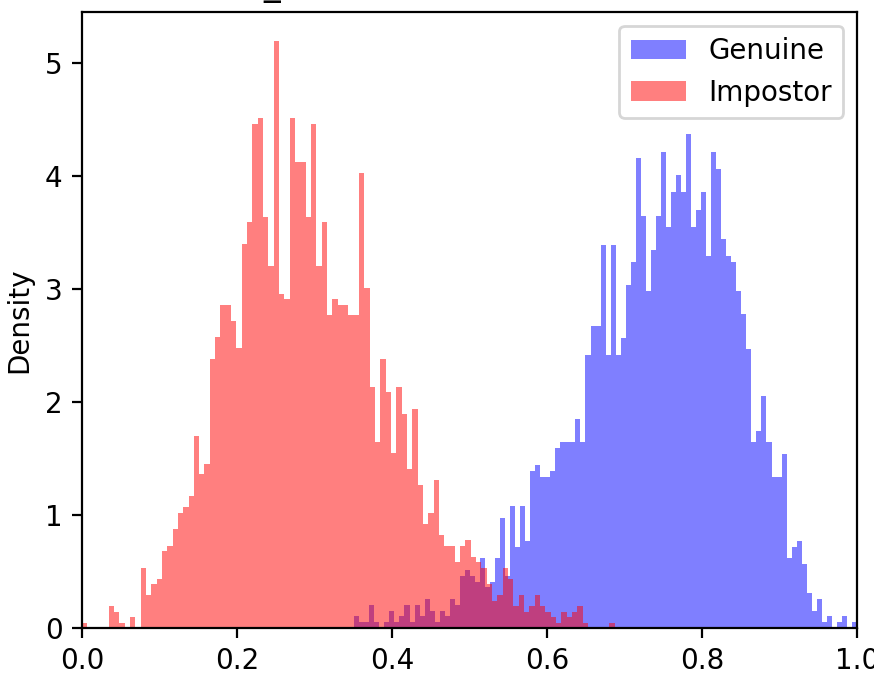}
        \caption*{LLaVA-1.5~\cite{llava15}, TMR: $91.73\%$}
    \end{subfigure}
    \hfill
    \begin{subfigure}[t]{0.32\textwidth}
        \centering
        \includegraphics[width=\linewidth]{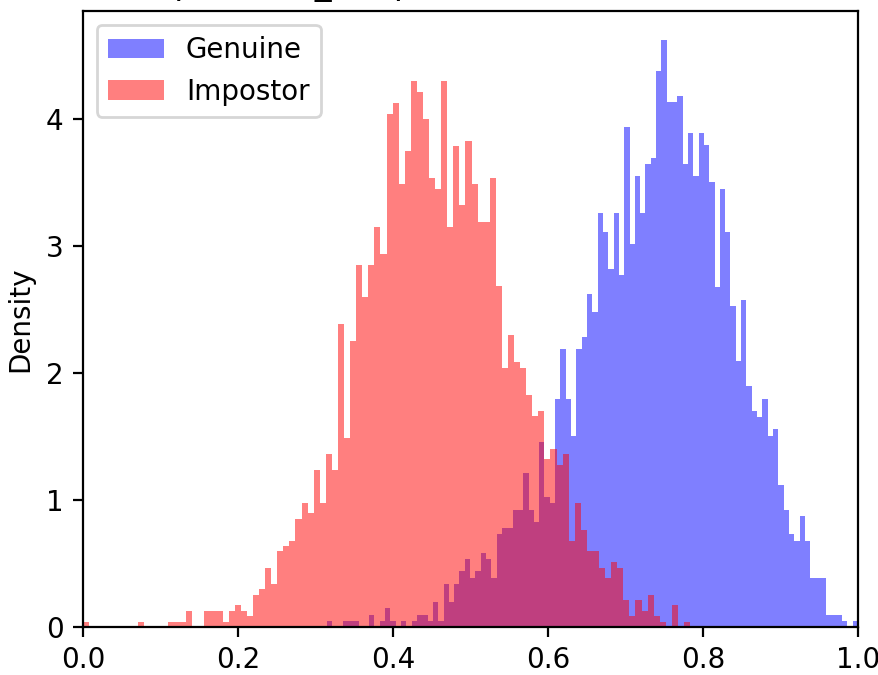}
        \caption*{DeepSeekVL2~\cite{wu2024deepseek_vl2}, TMR: $67.67\%$}
    \end{subfigure}
    \hfill
    \begin{subfigure}[t]{0.32\textwidth}
        \centering
        \includegraphics[width=\linewidth]{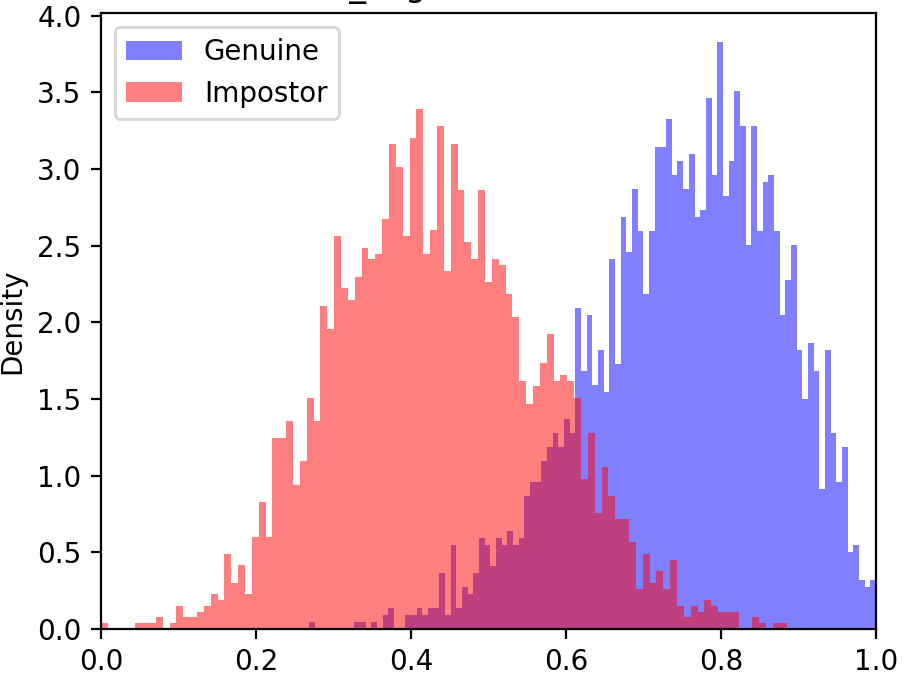}
        \caption*{ALIGN~\cite{jia2021scaling_align}, TMR: $50.67\%$}
    \end{subfigure}
    \vspace{0.5em}
    \caption{Score distributions of the genuine and impostor pairs in the LFW dataset~\cite{huang2008labeled} of the top six models for the face recognition task.}
    \label{fig:score_distribution_face}
\end{figure*}

\begin{figure*}[!ht]
    \centering
    \begin{subfigure}[t]{0.49\textwidth}
        \centering
        \fbox{\includegraphics[width=\linewidth]{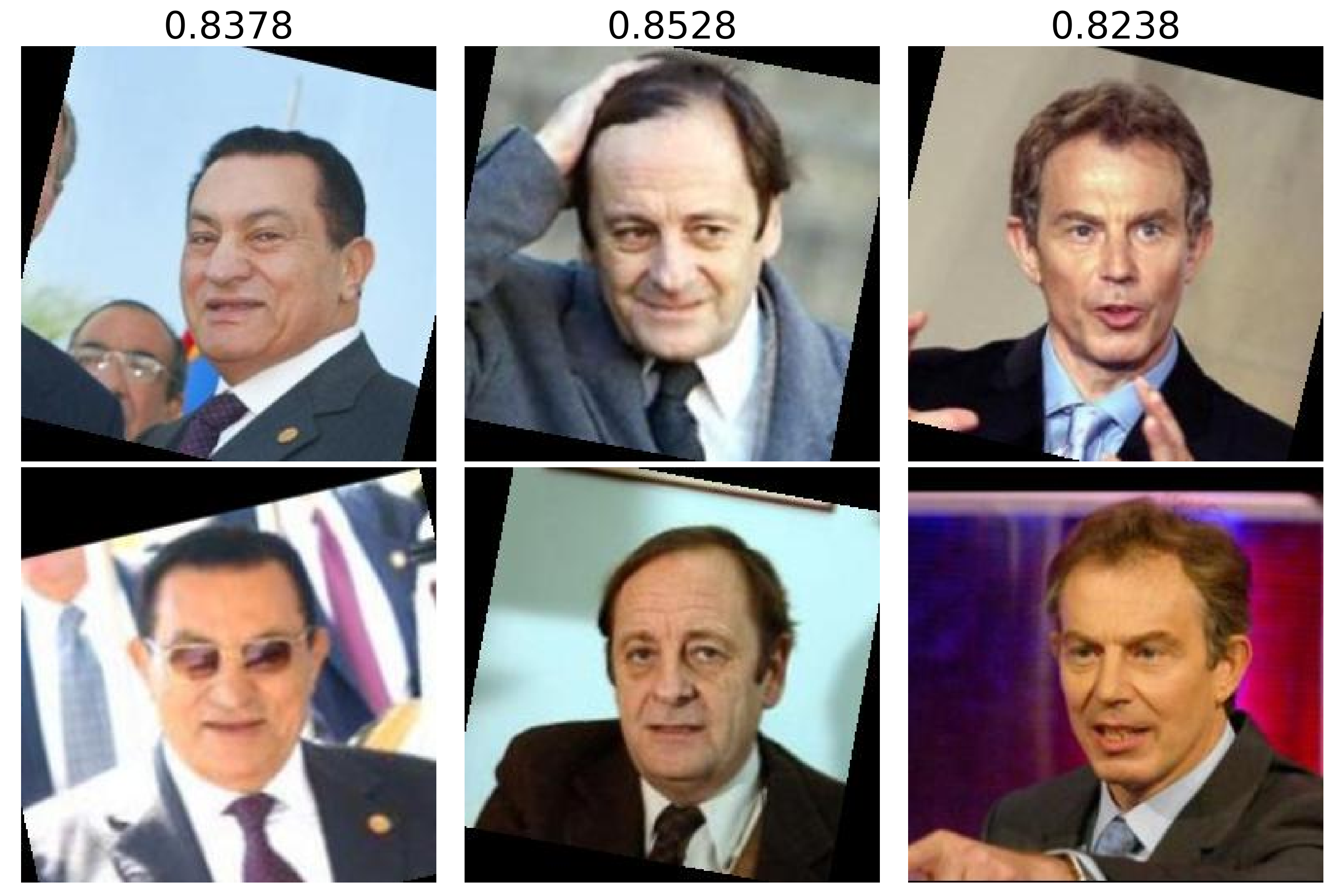}}
        \caption*{Genuine Pairs}
        \label{fig:full_iris}
    \end{subfigure}
    \hfill
    \begin{subfigure}[t]{0.49\textwidth}
        \centering
        \fbox{\includegraphics[width=\linewidth]{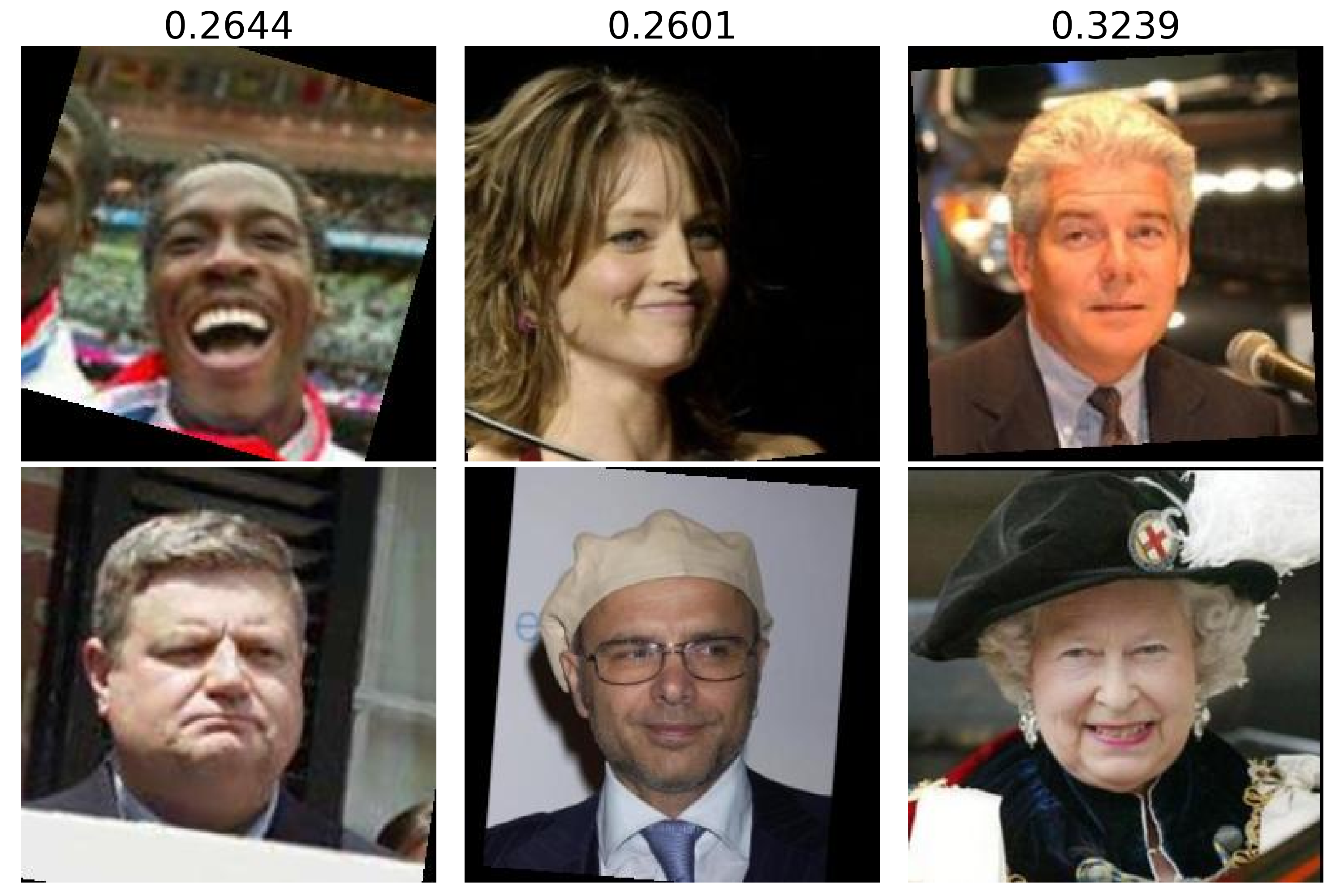}}
        \caption*{Impostor Pairs}
        \label{fig:cropped_iris}
    \end{subfigure}
\caption{Genuine (left three columns) and impostor (right three columns) pairs from the Labeled Faces in the Wild (LFW) dataset~\cite{huang2008labeled}. The match score (cosine similarity) from OpenCLIP-Huge~\cite{Cherti_2023_CVPR_OPENCLIP} model is displayed above each image pair.}
    \label{fig:lfw_samples}
\end{figure*}
For the AgeDB dataset, we follow the standard evaluation protocol and report \textbf{TMR@1\%FMR} across four age gap settings with increasing levels of difficulty. On the 5-year age gap protocol, \textbf{LLaVA-1.5}~\cite{llava15} achieves the highest performance with a TMR@1\%FMR of \textbf{64.47\%}. For the 10-year and 30-year protocols, \textbf{CLIP-L/14-336} (CLIP Large with patch size 14 and input size 336) performs best, achieving TMR of \textbf{54.03\%} and \textbf{34.87\%}, respectively. On the most challenging 40-year protocol, \textbf{LLaVA-1.5} again outperforms other models with a TMR@1\%FMR of \textbf{21.23\%}.




\begin{figure*}[!ht]
    \makebox[\textwidth][c]{%
        \includegraphics[width=1.03\linewidth]{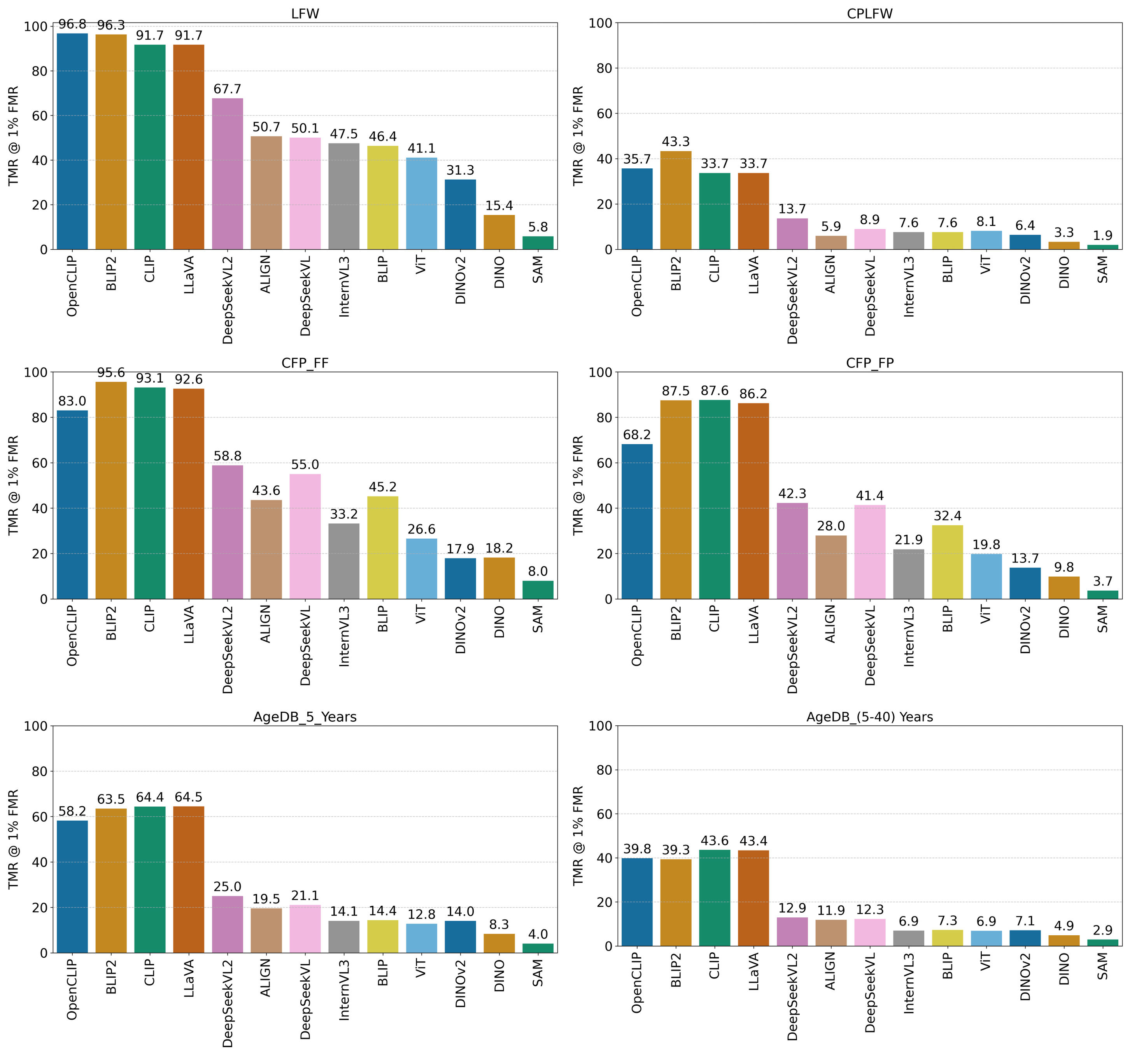}
    }
    \caption{Comparison of True Match Rate (TMR) at $1\%$ False Match Rate (FMR) for different face recognition (verification) tasks. For each model family, the model with best performance is reported.}
    \label{fig:combined_model_performance_face}
\end{figure*}
For the Labeled Faces in the Wild (LFW) dataset, the \textbf{OpenCLIP-H/14}~\cite{Cherti_2023_CVPR_OPENCLIP} model with patch size 14 achieved the highest score of \textbf{96.77\%}. However, \textbf{BLIP}~\cite{li2022blip}, \textbf{BLIP-2}~\cite{li2023blip2}, and \textbf{CLIP}~\cite{clip_radford2021learning} also demonstrated comparable performance. Figure~\ref{fig:score_distribution_face} shows the genuine and impostor socres distribution for the LFW benchmark. Figure~\ref{fig:lfw_samples} shows some genuine and impostor samples and their match scores from the best-performing model. On the more challenging Cross Pose LFW (\textbf{CPLFW}) dataset, the \textbf{BLIP}~\cite{li2022blip} model outperformed all others by a significant margin, highlighting its robustness under large pose variations.

On the Celebrity Frontal-Profile (\textbf{CFP})~\cite{cfp-paper_sengupta2016frontal} benchmark, for the Frontal-Frontal (\textbf{CFP\_FF}) setting, \textbf{BLIP}~\cite{li2022blip} and \textbf{BLIP-2}~\cite{li2023blip2} achieved strong results, with a top score of \textbf{95.54\%}. However, for the more difficult Frontal-Profile (\textbf{CFP\_FP}) protocol, the \textbf{CLIP-L/14}~\cite{lester2021powerscaleparameterefficientprompt} model (CLIP Large with patch size 14) achieved the best score of \textbf{87.63\%}, with \textbf{BLIP} and \textbf{BLIP-2} trailing closely behind.

Figure~\ref{fig:combined_model_performance_face} summarizes the combined performance of different VLMs for the face recognition task. These results are significant, as they demonstrate that foundation vision-language models can perform strongly on challenging biometric tasks like face recognition, despite not being explicitly trained for them. This underscores the effectiveness of language-guided unsupervised learning. Furthermore, some of these models, especially the smaller ones, achieve competitive accuracy while requiring significantly fewer parameters and computational resources.

\begin{table*}[!ht]
\centering
\caption{Ethnicity and gender classification accuracy on VMER~\cite{Greco_MVA2020_VMER} and AgeDB~\cite{moschoglou2017agedb} datasets respectively. The lightweight classifiers i.e., KNN~\cite{cover1967nearest}, LDA~\cite{fisher1936use_lda}, Logistic Regression~\cite{cox1958logiesticregression}, Ridge Regression~\cite{hoerl1970ridge}, and Support Vector Machine (SVM)~\cite{cortes1995svm} are trained on the features extracted by VLMs and then used for 5-Fold cross validation with disjoint set of identities in train and test split. Best accuracy for each method is bolded.}
\resizebox{1.00\textwidth}{!}{  
\begin{tabular}{l|ccccc||ccccc}
\Xhline{1.5px}
\multicolumn{11}{c}{\textbf{Attribute Prediction from Faces}}\\
\Xhline{1.5px}
\textbf{Model} & \multicolumn{5}{c||}{\textbf{Ethnicity Classification}} & \multicolumn{5}{c}{\textbf{Gender Classification}} \\
\cmidrule(lr){2-6} \cmidrule(lr){7-11}
 & \textbf{KNN }& \textbf{LDA} &\textbf{ Logistic} & \textbf{Ridge} & \textbf{SVM }&\textbf{ KNN} &\textbf{ LDA} & \textbf{Logistic }&\textbf{ Ridge} &\textbf{ SVM} \\
\midrule
InternVL3-1B~\cite{zhu2025internvl3}        & 82.90 & 94.54 & 94.10 & 94.30 & 93.16 & 98.02 & 99.78 & 99.77 & 99.80 & 99.75\\
InternVL3-38B~\cite{zhu2025internvl3}       & 84.64 & 89.30 & 93.84 & 92.96 & 93.64 & 96.60 & 99.82 & 99.78 & 99.84 & 99.75\\\hline
BLIP-Base~\cite{li2022blip}                 & 82.16 & 93.36 & 91.86 & 93.16 & 90.48 & 97.62 & 99.24 & 99.13 & 99.21 & 98.77\\
BLIP-Large~\cite{li2022blip}                & 84.76 & 93.86 & 92.60 & 93.96 & 90.96 & 99.27 & 99.65 & 99.56 & 99.64 & 99.48\\\hline
BLIP2-t5-xl~\cite{li2023blip2}              & 89.56 & 95.36 & 94.46 & 95.40 & 93.46 & 99.28 & 99.93 & 99.92 & \textbf{99.93} & 99.89\\
BLIP2-t5-xl-coco~\cite{li2023blip2}         & 89.80 & 95.32 & 94.36 & 95.44 & 93.82 & 99.54 & 99.90 & 99.88 & 99.90 & 99.82\\
BLIP2-t5-xxl~\cite{li2023blip2}             & 89.56 & 95.36 & 94.48 & 95.40 & 93.50 & 99.28 & 99.93 & 99.92 & 99.93 & 99.89\\
BLIP2-O-2.7b~\cite{li2023blip2}             & 89.56 & 95.32 & 94.46 & 95.40 & 93.50 & 99.28 & 99.93 & 99.92 & 99.93 & 99.89\\
BLIP2-O-6.7b~\cite{li2023blip2}             & 89.56 & 95.34 & 94.46 & 95.40 & 93.52 & 99.27 & 99.93 & 99.92 & \textbf{99.93} & 99.89\\\hline
DeepSeekVL-1.3B~\cite{lu2024deepseek_vl}    & 89.98 & 94.94 & 94.70 & 94.82 & 94.26 & 99.22 & 99.80 & 99.78 & 99.80 & 99.74\\
DeepSeekVL-7B~\cite{lu2024deepseek_vl}      & 89.94 & 95.10 & 94.74 & 94.90 & 94.40 & 99.25 & 99.80 & 99.78 & 99.78 & 99.75\\\hline
DeepSeek-VL2~\cite{dai2024deepseekmoe}      & 87.12 & 95.36 & 91.56 & 88.10 & 95.30 & 98.93 & 99.83 & 99.82 & 99.86 & 99.86\\
DeepSeek-VL2-S~\cite{dai2024deepseekmoe}    & 86.34 & 95.60 & 93.98 & 92.92 & \textbf{95.84} & 98.90 & 99.85 & 99.85 & 99.88 & 99.90\\
DeepSeek-VL2-T~\cite{dai2024deepseekmoe}    & 86.28 & 95.10 & 93.84 & 92.70 & 95.46 & 98.84 & 99.82 & 99.84 & 99.87 & 99.88\\\hline
DINO-ViTb16~\cite{caron2021dino}            & 72.10 & 82.74 & 81.10 & 83.32 & 79.52 & 89.69 & 95.87 & 93.59 & 95.83 & 92.83\\\hline
DINOv2-Base~\cite{oquab2023dinov2}          & 79.90 & 89.96 & 88.06 & 89.76 & 87.14 & 97.55 & 98.87 & 98.65 & 98.87 & 98.51\\
DINOv2-Giant~\cite{oquab2023dinov2}         & 80.20 & 89.96 & 90.52 & 89.84 & 90.18 & 97.55 & 99.33 & 99.19 & 99.34 & 99.07\\
DINOv2-Large~\cite{oquab2023dinov2}         & 80.40 & 90.62 & 89.12 & 90.42 & 88.42 & 97.57 & 99.25 & 99.20 & 99.26 & 99.05\\
DINOv2-Small~\cite{oquab2023dinov2}         & 76.72 & 87.28 & 84.16 & 86.54 & 81.96 & 95.75 & 98.11 & 97.22 & 98.11 & 96.85\\\hline
SAM-ViT-Base~\cite{kirillov2023segment_anything}    & 63.06 & 72.14 & 76.00 & 76.00 & 76.00 & 62.68 & 81.69 & 71.52 & 70.58 & 72.25\\
SAM-ViT-Huge~\cite{kirillov2023segment_anything}    & 62.44 & 73.76 & 76.00 & 76.00 & 76.00 & 65.29 & 85.63 & 72.73 & 71.05 & 73.66\\
SAM-ViT-Large~\cite{kirillov2023segment_anything}   & 63.02 & 72.92 & 76.00 & 76.00 & 76.00 & 64.68 & 83.70 & 73.73 & 72.37 & 74.78\\\hline
ViT-B-16~\cite{dosovitskiy2020image_vit}            & 84.90 & 91.76 & 89.16 & 91.36 & 88.64 & 96.65 & 98.27 & 97.52 & 98.27 & 97.31\\
ViT-B-16-384~\cite{dosovitskiy2020image_vit}        & 86.58 & 92.14 & 89.44 & 91.80 & 89.08 & 97.39 & 98.89 & 98.41 & 98.89 & 98.19\\
ViT-B-32-384~\cite{dosovitskiy2020image_vit}        & 85.32 & 90.54 & 87.50 & 90.76 & 87.56 & 96.12 & 97.92 & 96.96 & 97.92 & 96.76\\
ViT-H-14-224-21k~\cite{dosovitskiy2020image_vit}    & 84.34 & 88.94 & 87.24 & 88.74 & 87.06 & 95.22 & 97.14 & 96.13 & 97.14 & 95.80\\
ViT-Hybrid-384~\cite{dosovitskiy2020image_vit}      & 85.44 & 92.50 & 90.74 & 92.24 & 90.00 & 97.11 & 98.72 & 98.72 & 98.75 & 98.49\\
ViT-L-16-224~\cite{dosovitskiy2020image_vit}        & 85.16 & 91.82 & 90.40 & 91.16 & 89.96 & 96.54 & 98.39 & 97.94 & 98.39 & 97.64\\
ViT-L-16-384~\cite{dosovitskiy2020image_vit}        & 85.54 & 91.82 & 91.06 & 91.76 & 90.48 & 97.41 & 98.81 & 98.60 & 98.81 & 98.40\\
ViT-L-32-384~\cite{dosovitskiy2020image_vit}        & 83.14 & 90.60 & 88.26 & 90.52 & 88.60 & 96.40 & 98.40 & 97.75 & 98.40 & 97.68\\\hline
OpenCLIP-B-32~\cite{Cherti_2023_CVPR_OPENCLIP}      & 90.22 & 94.56 & 94.14 & 93.98 & 94.26 & 99.11 & 99.43 & 99.46 & 99.46 & 99.51\\
OpenCLIP-H-14~\cite{Cherti_2023_CVPR_OPENCLIP}      & 90.62 & 95.70 & 94.54 & 94.96 & 95.04 & 99.33 & 99.91 & 99.93 & \textbf{99.93} & \textbf{99.92}\\
OpenCLIP-G-14~\cite{Cherti_2023_CVPR_OPENCLIP}      & 90.22 & 95.56 & \textbf{94.76} & 95.00 & 95.18 & 99.45 & 99.91 & 99.93 & 99.91 & 99.88\\\hline
ALIGN~\cite{jia2021scaling_align}                   & 89.76 & 95.62 & 94.28 & 95.34 & 93.38 & 99.20 & 99.62 & 99.64 & 99.63 & 99.42\\\hline
LLaVA-1.5-7B~\cite{llava15}                         & 89.58 & 95.32 & 93.60 & 95.04 & 93.26 & 98.74 & 99.93 & 99.93 & \textbf{99.93} & \textbf{99.92}\\
LLaVA-1.6-M-7b~\cite{liu2024llavanext}              & 82.76 & 50.78 & 94.32 & 94.36 & 93.74 & 97.23 & 99.81 & 99.76 & 99.77 & 99.79\\\hline
CLIP-B-16~\cite{clip_radford2021learning}           & \textbf{91.50} & \textbf{95.78} & 94.46 & \textbf{95.56 }& 93.94 & \textbf{99.63} & 99.89 & 99.89 & 99.88 & 99.89\\
CLIP-B-32~\cite{clip_radford2021learning}           & 90.62 & 95.12 & 94.46 & 95.26 & 93.64 & 99.36 & 99.73 & 99.68 & 99.72 & 99.66\\
CLIP-L-32~\cite{clip_radford2021learning}           & 89.06 & 95.38 & 93.72 & 94.70 & 93.54 & 98.55 & 99.93 & \textbf{99.94} & \textbf{99.93} & \textbf{99.92}\\
CLIP-L-14-336~\cite{clip_radford2021learning}       & 89.56 & 95.30 & 93.60 & 95.02 & 93.28 & 98.75 & \textbf{99.93} & 99.93 & \textbf{99.93} & \textbf{99.92}\\
\bottomrule
\end{tabular}%
}
\label{table:age_gender_combined}
\end{table*}
\subsection{Gender Classification}

We evaluated foundation models for their performance in soft biometric feature extraction, with a focus on gender classification. To this end, we used a version of the \textbf{AgeDB}~\cite{moschoglou2017agedb} dataset—a face recognition benchmark that includes gender annotations. Details of the dataset can be found in Section~\ref{sec:agedb_dataset}. We first extracted feature vectors for each face image using various vision-language models (VLMs). Since the dimensionality of the extracted features varies across models, we trained lightweight classifiers on top of these features to assess performance in terms of classification accuracy. The results, averaged over 5-fold cross-validation with disjoint identities in train and test splits, are reported in Table~\ref{table:age_gender_combined}.
Our findings indicate that most models i.e., CLIP, BLIP, OpenCLIP and LLaVa retain gender information reasonably well (most of them are around $99.9\%$ accurate). An exception is the Segment Anything Model (\textbf{SAM})~\cite{kirillov2023segment_anything}, which performs poorly on this task. This is expected, as SAM is primarily designed for segmentation and object-level understanding rather than identity or attribute-level feature representation.

\subsection{Ethnicity Classification}
As an important soft biometric attribute, we evaluate the performance of foundation vision-language models on the task of ethnicity classification. For this purpose, we utilize the VGG-Face2 Mivia Ethnicity Recognition (VMER) dataset~\cite{Greco_MVA2020_VMER}, which includes annotations for four distinct ethnicity classes on a subset of the data. 
Given the large number of samples per identity in the original dataset, we subsample 10 images per identity to preserve class balance while maintaining a manageable size. We extract face embeddings using the vision encoders of various foundation models and train lightweight shallow classifiers on top of these features. Classification performance is evaluated using 5-fold cross-validation with a disjoint set of identities in train and testing. The Table~\ref{table:age_gender_combined} presents the average classification accuracies obtained across different models. Among the evaluated models, small variant of \textbf{DeepSeekVL2}~\cite{wu2024deepseek_vl2} showed the best performance with an SVM classification head with an accuracy of \textbf{$95.84\%$}. Among other models, \textbf{LLaVA-1.5}~\cite{llava15}, \textbf{CLIP-L/14-336}~\cite{clip_radford2021learning}, \textbf{OpenCLIP}~\cite{Cherti_2023_CVPR_OPENCLIP}, \textbf{DeepSeekVL}~\cite{lu2024deepseek_vl}, and \textbf{BLIP}~\cite{li2022blip, li2023blip2} also demonstrated competitive performance.

\subsection{Iris Recognition}
\begin{table}[!ht]
\centering
\caption{\textbf{TMR@1\%FMR }of vision–language models on four iris recognition benchmarks: UND-Full, UND-Cropped, IITD-R-Full, and IITD-R-Cropped.}
\label{tab:iris-recognition-vlm}
\resizebox{1.00\linewidth}{!}{  
\begin{tabular}{l|rr||rc}
\Xhline{1.5px}
\multicolumn{5}{c}{\textbf{Iris Recognition Task}} \\
\Xhline{1.5px}
& \multicolumn{2}{c||}{\textbf{UND~\cite{bowyer2008image_und_iris}}} & \multicolumn{2}{c}{\textbf{IITD-R ~\cite{kumar2010comparison}}} \\
\cmidrule(lr){2-3} \cmidrule(lr){4-5}
\textbf{Model} & \textbf{Full} & \textbf{Crop.} & \textbf{Full} & \textbf{Crop.} \\

\Xhline{1.5px}
Kosmos~\cite{peng2023kosmos} & 9.09 & 9.32 & 63.37 & 49.52 \\\hline
BLIP-Base~\cite{li2022blip} & 6.69 & 12.26 & 64.68 & 59.05 \\
BLIP-Large~\cite{li2022blip} & 6.92 & 8.24 & 41.37 & 40.18 \\\hline
BLIP2-t5-xl~\cite{li2023blip2} & 4.12 & 6.27 & 30.18 & 28.76 \\
BLIP2-t5-xl-coco~\cite{li2023blip2} & 4.12 & 6.27 & 26.75 & 22.72 \\
BLIP2-t5-xxl~\cite{li2023blip2} & 4.12 & 6.27 & 34.42 & 28.10 \\
BLIP2-opt-2.7b~\cite{li2023blip2} & 4.12 & 6.27 & 33.01 & 31.45 \\
BLIP2-opt-6.7b~\cite{li2023blip2} & 4.12 & 6.27 & 34.03 & 29.41 \\\hline
DeepSeekVL-1.3B-B~\cite{lu2024deepseek_vl} & 23.73 & 9.41 & 89.60 & 47.87 \\
DeepSeekVL-1.3B-C~\cite{lu2024deepseek_vl} & 24.56 & 10.07 & 91.12 & 56.32 \\
DeepSeekVL-7B-B~\cite{lu2024deepseek_vl} & 24.02 & 9.91 & 90.18 & 49.48 \\
DeepSeekVL-7B-C~\cite{lu2024deepseek_vl} & 24.12 & 10.18 & 90.16 & 53.29 \\\hline
DeepSeekVL2~\cite{wu2024deepseek_vl2} & 35.57 & 39.22 & 92.91 & 86.66 \\
DeepSeekVL2-S~\cite{wu2024deepseek_vl2} & 25.33 & 39.64 & 90.41 & 86.04 \\
DeepSeekVL2-T~\cite{wu2024deepseek_vl2} & 27.65 & 33.01 & 92.60 & 83.74 \\\hline
DINO-ViTB16~\cite{caron2021dino} & \textbf{42.10} & \textbf{51.67} & \textbf{97.55} & \textbf{96.66} \\
DINOv2-Base~\cite{oquab2023dinov2} & 21.69 & 19.97 & 73.07 & 72.24 \\
DINOv2-Giant~\cite{oquab2023dinov2} & 24.11 & 24.05 & 76.64 & 68.40 \\
DINOv2-Large~\cite{oquab2023dinov2} & 22.79 & 18.57 & 75.79 & 70.86 \\
DINOv2-Small~\cite{oquab2023dinov2} & 23.57 & 23.41 & 82.20 & 86.08 \\\hline
SAM-ViT-Base~\cite{kirillov2023segment_anything} & 10.41 & 19.91 & 81.86 & 73.51 \\
SAM-ViT-Huge~\cite{kirillov2023segment_anything} & 10.49 & 16.93 & 74.15 & 65.86 \\
SAM-ViT-Large~\cite{kirillov2023segment_anything} & 13.63 & 17.74 & 72.86 & 65.21 \\\hline
OpenCLIP-B-32~\cite{Cherti_2023_CVPR_OPENCLIP} & 12.18 & 13.11 & 59.61 & 57.02 \\
OpenCLIP-H-14~\cite{Cherti_2023_CVPR_OPENCLIP} & 15.87 & 16.37 & 74.88 & 53.13 \\
OpenCLIP-G-14~\cite{Cherti_2023_CVPR_OPENCLIP} & 14.38 & 19.22 & 81.84 & 63.63 \\\hline
LLaVA-1.5-7B~\cite{llava15} & 17.25 & 14.96 & 63.23 & 47.52 \\
LLaVA-1.6-M-7b~\cite{liu2024llavanext} & 15.85 & 18.58 & 68.22 & 45.67 \\\hline
CLIP-B-16~\cite{clip_radford2021learning} & 14.06 & 14.67 & 52.39 & 53.17 \\
CLIP-B-32~\cite{clip_radford2021learning} & 11.48 & 16.19 & 65.25 & 60.09 \\
CLIP-L-14~\cite{clip_radford2021learning} & 13.28 & 16.76 & 46.15 & 50.52 \\
CLIP-L-14-336~\cite{clip_radford2021learning} & 14.75 & 16.94 & 63.64 & 48.40 \\
\Xhline{1.5px}
\end{tabular}}

\end{table}
\begin{figure*}[!ht]
    \centering
    \makebox[\textwidth][c]{%
        \includegraphics[width=1.00\textwidth]{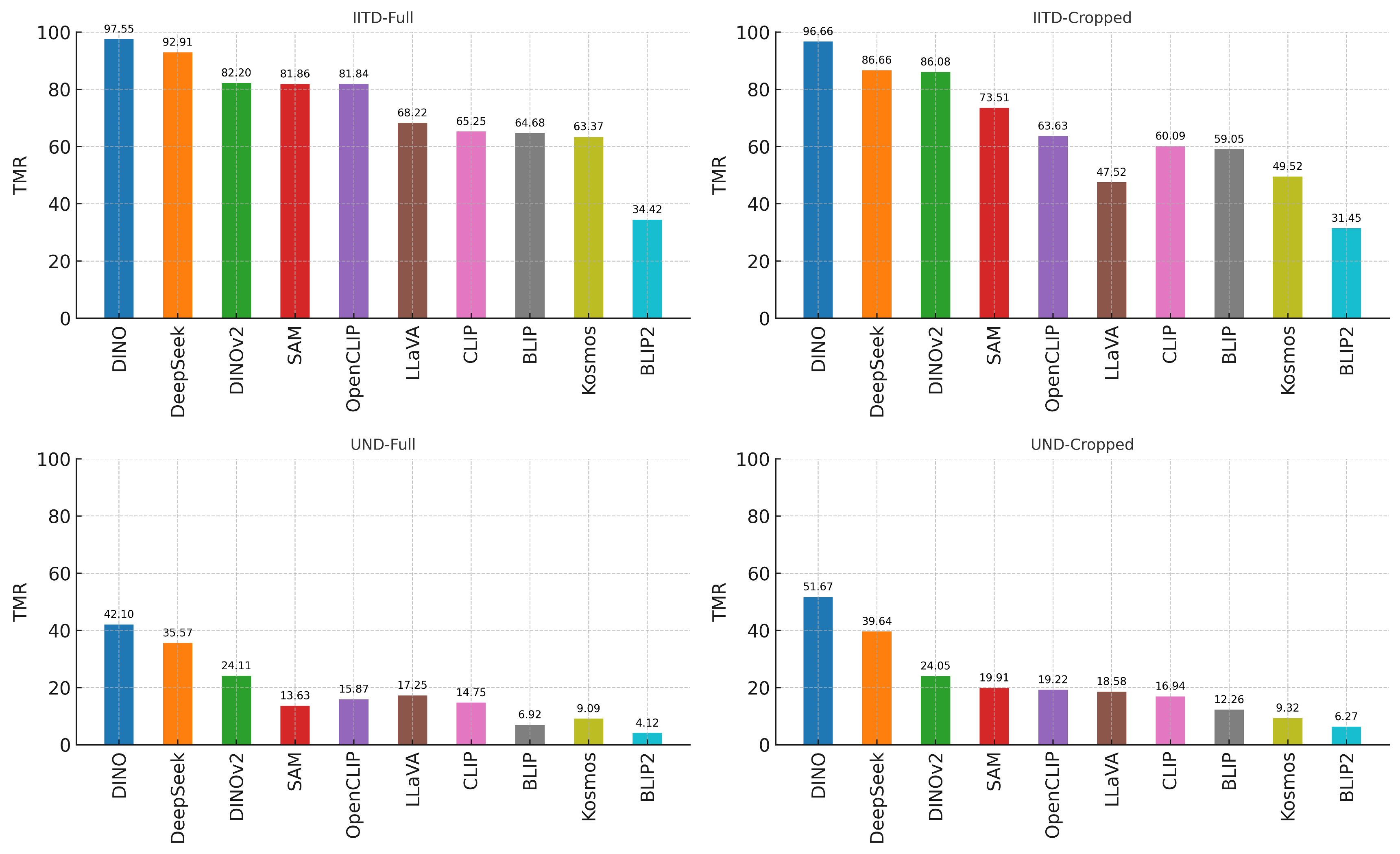}
    }
    \caption{Comparison of TMR@1\%FMR across 10 vision-language model groups on four iris recognition datasets: UND-Full~\cite{bowyer2008image_und_iris}, UND-Cropped~\cite{bowyer2008image_und_iris}, IITD-R-Full~\cite{kumar2010comparison}, and IITD-R-Cropped~\cite{kumar2010comparison}. For each model group, the best-performing configuration is selected.}
    \label{fig:auc_tmr_IrisRec}
\end{figure*}

Table~\ref{tab:iris-recognition-vlm} and Figure~\ref{fig:auc_tmr_IrisRec} summarize the zero-shot iris recognition performance of various vision--language models (VLMs) evaluated on the UND-Iris-0405~\cite{bowyer2008image_und_iris} and IIT-Delhi-Iris (IITD-R)~\cite{kumar2010comparison} datasets. Both full and cropped versions of the iris images are used. Performance is measured in terms of True Match Rate at 1\% False Match Rate (TMR@1\%FMR). To obtain these scores, we first extracted deep visual features from each model’s vision encoder in a zero-shot manner, without any fine-tuning or task-specific supervision. The extracted features were then used to compute pairwise cosine similarity scores across iris image pairs. Labels were determined by identity information in filenames, and performance was evaluated using ground-truth labels to compute TMR.

Across all benchmarks, the \textbf{DINO}~\cite{caron2021dino} and \textbf{DINOv2}~\cite{oquab2023dinov2} families consistently outperform other models on the IITD-R dataset. In particular, \textbf{DINO} achieves top-tier performance across all four dataset splits using the TMR metric. Additionally, \textbf{DINOv2-Small} shows the second-highest scores on the IITD-R dataset, followed closely by other \textbf{DINOv2} variants including \textbf{Base}, \textbf{Giant}, and \textbf{Large}. These findings suggest that the self-supervised vision encoders in the \textbf{DINOv2} series are especially well-suited for iris recognition tasks—potentially due to their training on diverse, high-resolution datasets or implicit exposure to ocular patterns during pretraining.
Importantly, we observe that the IITD-R dataset yields higher TMR scores compared to the UND dataset for nearly all models. This trend is consistent across \textbf{DINOv2~\cite{oquab2023dinov2}, SAM~\cite{kirillov2023segment_anything}, CLIP~\cite{clip_radford2021learning}, OpenCLIP~\cite{Cherti_2023_CVPR_OPENCLIP}, DeepSeek~\cite{lu2024deepseek_vl}}, and others, indicating that the IITD-R dataset may be less challenging in the zero-shot setting or more visually consistent with the pretraining distribution of these models.

Interestingly, there is no substantial difference in performance between full and cropped iris images for most models. This suggests that many of these vision encoders are able to focus on relevant iris features even when unsegmented regions such as eyelids and sclera are present.

At the lower end of the spectrum, the BLIP2 variants exhibit the weakest performance across all datasets, with TMR values as low as 4--6\% on the UND~\cite{bowyer2008image_und_iris} dataset. However, while the performance on the IITD-R~\cite{iitd_iris} dataset improved, it remains the lowest among all evaluated models. These results highlight the limitations of certain VLMs that are optimized for multimodal alignment tasks but may not produce discriminative enough features for fine-grained iris recognition. Additionally, among all models evaluated on the UND-Full dataset, the \textbf{DINO}~\cite{caron2021dino} model achieves the highest performance, indicating its relative robustness on this benchmark compared to other VLMs.

Overall, these findings demonstrate that zero-shot models such as \textbf{DINO/DINOv2} show potential for application in iris recognition. However, performance still varies significantly depending on model architecture, training strategy, and dataset domain, underscoring the need for further exploration into domain adaptation and iris-specific fine-tuning.

\begin{figure*}[!ht]
  \centering
  \includegraphics[width=\linewidth]{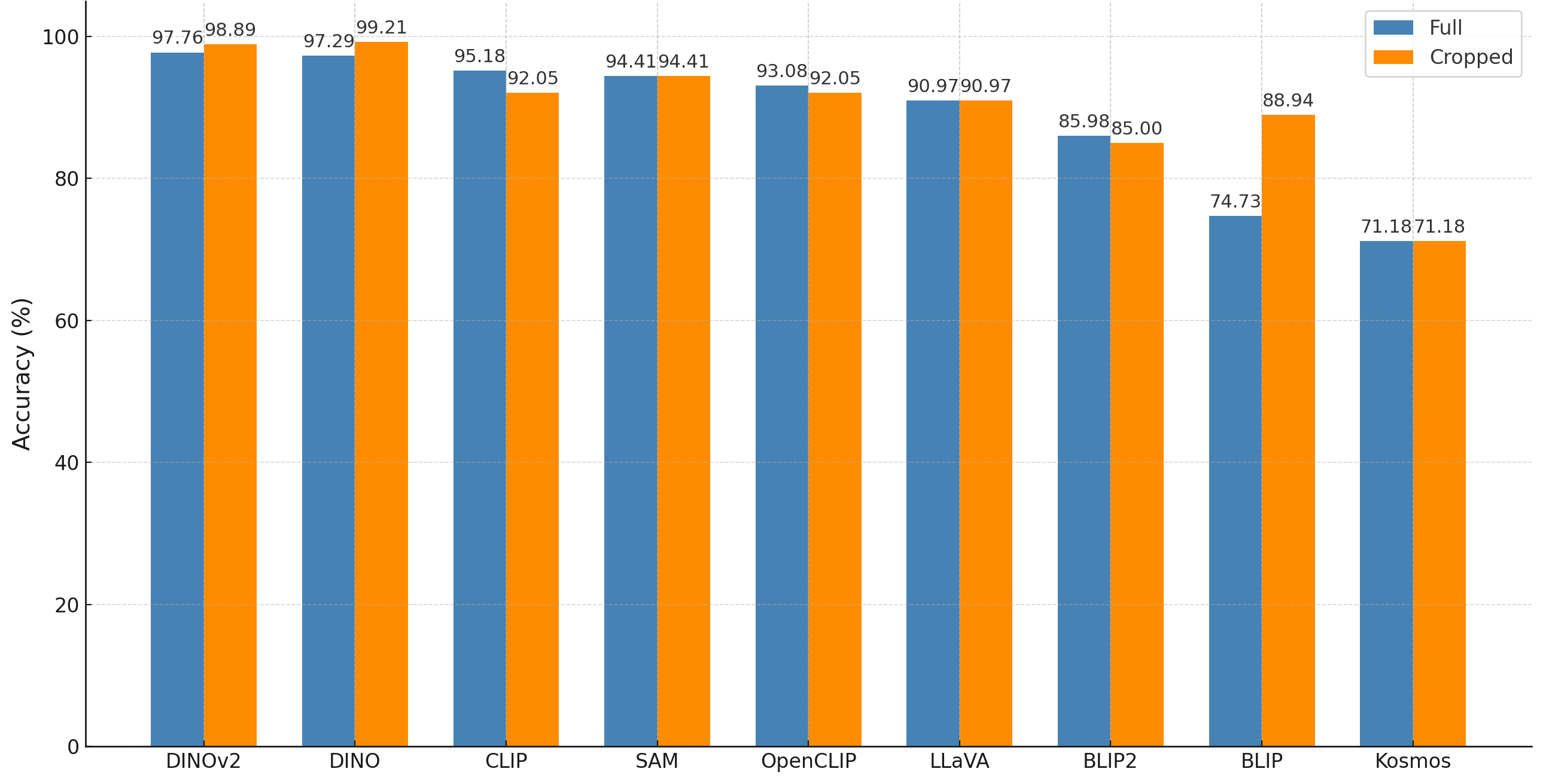}  
  \caption{Accuracy of different VLMs on the IIIT-Delhi-Iris (IITD-P) dataset~\cite{iitd_iris} for Presentation Attack (PA) detection task. The best performing model from each family is shown for both full and cropped iris images.}
  \label{fig:iitdpa_bar}
\end{figure*}

\begin{table*}[!ht]
\centering
\captionsetup{width=\linewidth}            
\caption{IITD-P PA benchmark. Each model is evaluated on full and cropped images.
For each view we report one overall accuracy (Acc) and the accuracies of three classifiers
(SVM-lin, SVM-rbf, Logistic Regression).}
\label{tab:iitd-pa-table}

\setlength{\tabcolsep}{10pt}
\resizebox{1.00\textwidth}{!}{  
\begin{tabular}{l|cccc|cccc}
\Xhline{1.5px}
\multicolumn{9}{c}{\textbf{Iris Presentation Attack Detection Task}} \\
\Xhline{1.5px}
\textbf{Model} &
\multicolumn{4}{c|}{\textbf{IITD-P Full}~\cite{singh2018ghclnet}} &
\multicolumn{4}{c}{\textbf{IITD-P Cropped}~\cite{singh2018ghclnet}} \\ 
\cmidrule(lr){2-5} \cmidrule(lr){6-9}
~ & \textbf{Acc} & \textbf{SVM} & \textbf{RBF} & \textbf{Log} &
    \textbf{Acc} & \textbf{SVM} & \textbf{RBF} & \textbf{Log} \\
\Xhline{1.5px}
Kosmos~\cite{peng2023kosmos}                             & 71.18 & 98.08 & 89.96 & 97.86 & 71.18 & 98.29 & 90.60 & 97.86 \\ \hline
BLIP-Base~\cite{li2022blip}                  & 74.73 & 89.74 & 78.21 & 88.46 & 87.11 & 98.06 & 93.12 & 98.06 \\
BLIP-Large~\cite{li2022blip}                 & 73.44 & 88.89 & 77.35 & 87.82 & 88.94 & 98.28 & 95.48 & 98.28 \\\hline
BLIP2-t5-xl~\cite{li2023blip2}          & 85.96 & \textbf{99.57} & 94.66 & 99.15 & 84.94 & 99.35 & 98.28 & 98.92 \\
BLIP2-t5-xl-coco~\cite{li2023blip2}     & 75.00 & 97.44 & 91.67 & 97.01 & 84.94 & 99.35 & 98.28 & 98.92 \\
BLIP2-t5-xxl~\cite{li2023blip2}         & 85.98 & \textbf{99.57} & 94.66 & 99.15 & 85.00 & 99.35 & 98.28 & 98.92 \\
BLIP2-O-2.7b~\cite{li2023blip2}            & 85.97 & \textbf{99.57} & 94.44 & 98.93 & 84.97 & 99.35 & 98.28 & 98.92 \\
BLIP2-O-6.7b~\cite{li2023blip2}            & 85.92 & \textbf{99.57} & 94.66 & 98.93 & 84.96 & 99.35 & 98.28 & 98.92 \\ \hline
DINO-ViTB16~\cite{caron2021dino}            & 97.29 & 99.19 & 99.57 & 99.49 & \textbf{99.21} & \textbf{99.84} & \textbf{99.89} & \textbf{99.87} \\
DINOv2-Base~\cite{oquab2023dinov2}           & \textbf{97.76} & 99.43 & \textbf{99.65} & \textbf{99.52} & 98.21 & 98.57 & 97.95 & 98.36 \\
DINOv2-Giant~\cite{oquab2023dinov2}          & 94.73 & 98.78 & 98.78 & 98.78 & 98.89 & 98.36 & 98.77 & 98.77 \\
DINOv2-Large~\cite{oquab2023dinov2}          & 96.16 & 98.37 & 97.35 & 98.17 & 96.69 & 98.98 & 98.98 & 98.77 \\
DINOv2-Small~\cite{oquab2023dinov2}          & 90.96 & 95.52 & 94.09 & 97.15 & 98.00 & 98.57 & 97.54 & 98.77 \\ \hline
SAM-ViT-Base~\cite{kirillov2023segment_anything} & 94.41 & 82.58 & 86.89 & 83.40 & 94.41 & 82.58 & 86.89 & 83.40 \\
SAM-ViT-Huge~\cite{kirillov2023segment_anything} & 89.64 & 84.63 & 86.89 & 84.02 & 89.64 & 84.63 & 86.89 & 84.02 \\
SAM-ViT-Large~\cite{kirillov2023segment_anything}& 89.64 & 84.63 & 86.89 & 84.02 & 89.64 & 84.63 & 86.89 & 84.02 \\ \hline
OpenCLIP-B-32~\cite{Cherti_2023_CVPR_OPENCLIP} & 90.27 & 93.28 & 92.67 & 93.69 & 92.05 & 95.49 & 92.83 & 94.67 \\
OpenCLIP-H-14~\cite{Cherti_2023_CVPR_OPENCLIP} & 90.27 & 93.28 & 92.67 & 93.69 & 92.05 & 95.49 & 92.83 & 94.67 \\
OpenCLIP-G-14~\cite{Cherti_2023_CVPR_OPENCLIP} & 93.08 & 97.35 & 93.89 & 96.13 & 92.05 & 95.49 & 92.83 & 94.67 \\ \hline
LLaVA-1.5-7B~\cite{llava15}               & 90.97 & 97.96 & 94.30 & 97.76 & 90.97 & 98.17 & 94.30 & 96.76 \\
LLaVA-1.6-M-7b~\cite{liu2024llavanext}       & 90.97 & 98.17 & 94.50 & 97.76 & 90.97 & 97.96 & 94.30 & 97.76 \\ \hline
CLIP-B-16~\cite{clip_radford2021learning}     & 71.97 & 93.69 & 84.32 & 93.89 & 71.11 & 94.47 & 87.30 & 93.44 \\
CLIP-B-32~\cite{clip_radford2021learning}     & 90.27 & 93.28 & 92.67 & 93.69 & 92.05 & 95.49 & 92.83 & 94.67 \\
CLIP-L-14~\cite{clip_radford2021learning}     & 95.18 & 97.15 & 96.74 & 98.17 & 85.28 & 97.13 & 95.29 & 97.54 \\
CLIP-L-14-336~\cite{clip_radford2021learning} & 87.51 & 97.15 & 90.43 & 97.96 & 84.40 & 98.57 & 91.39 & 98.16 \\
\Xhline{1.5px}
\end{tabular}}
\end{table*}

\subsection{Iris Presentation Attack Detection}
Table~\ref{tab:iitd-pa-table} presents the results of iris presentation attack detection (PAD) using a suite of vision–language models (VLMs) on the IITD-P dataset~\cite{singh2018ghclnet}. The task focuses on binary classification between \textbf{bona fide (Normal)} and \textbf{presentation attack (Patterned)} iris images. We evaluated performance across two settings: using the full eye region and cropped iris-only views.

In the first stage, we conducted zero-shot inference. The resulting classification accuracy is reported under the \textbf{Acc} column, reflecting each model's raw discriminative capability in the absence of supervised training. To evaluate whether simple downstream classifiers could improve upon the zero-shot performance, we trained three standard classifiers on the extracted features: a linear Support Vector Machine (SVM), an SVM with radial basis function kernel (SVM-RBF), and Logistic Regression (Log). An identity-aware 80/20 subject-disjoint split was used to train and test the classifiers, ensuring that no identity appeared in both sets. The accuracy of each trained classifier is reported under the columns \textbf{SVM}, \textbf{RBF}, and \textbf{Log} in Table~\ref{tab:iitd-pa-table}.

Overall, classifier training significantly boosts performance over the zero-shot baseline, with many models achieving near-perfect accuracy. Among all models, the \textbf{DINOv2} variants consistently achieve the highest performance, with \textbf{DINO-ViTB16} and \textbf{DINOv2-Base} attaining accuracies exceeding \textbf{97\%} across both ocular and cropped images. This suggests that DINOv2's feature representations are particularly well aligned with the PAD task, possibly due to their training on diverse and fine-grained visual data. The \textbf{BLIP2} variants also demonstrated strong results when using a \textbf{linear SVM} on the ocular iris images. Specifically, \texttt{BLIP2-t5-xl}, \texttt{BLIP2-t5-xxl}, and \texttt{BLIP2-O-6.7b} models achieved over \textbf{99\%} SVM accuracy on both ocular and cropped IITD-P data. This suggests that although BLIP2 embeddings may not be as discriminative in the zero-shot setup (Acc column), they become highly separable with linear decision boundaries when even minimal supervised fine-tuning is applied. 

These findings indicate that BLIP2 embeddings encode a feature space where classes (Patterned vs. Normal) are more linearly separable, making them particularly amenable to simple classification methods like linear SVMs---despite not being the most powerful in raw zero-shot performance. This highlights the nuanced interplay between feature representation geometry and classifier type in presentation attack detection tasks.

Other top-performing models include \textbf{OpenCLIP}, \textbf{LLaVA}, and \textbf{CLIP}, all of which demonstrate strong generalization when paired with simple classifiers. While cropped images often yield marginally higher scores, the improvement is less pronounced than in iris recognition, indicating that presentation attacks can be detected reliably even in the presence of surrounding ocular context.

These results demonstrate that pretrained VLMs offer a robust foundation for PAD tasks when paired with lightweight classifiers, even in the absence of end-to-end fine-tuning. Figure~\ref{fig:iitdpa_bar} presents the performance of all models, ranked from highest to lowest accuracy.

\subsection{DeepFake Detection}
\begin{table}[!ht]
\centering
\caption{DeepFake detection accuracy of the simple classifier head when trained on the feature embedding extracted from different vision language models.}
\resizebox{1.00\linewidth}{!}{  
\begin{tabular}{l|c|c|c}
\Xhline{1.5px}
\multicolumn{4}{c}{\textbf{DeepFake Detection (Face)}}\\
\Xhline{1.5px}
\textbf{Model} & \textbf{MLP} & \multicolumn{2}{c}{\textbf{Logistic Regression}} \\\cline{3-4}
 & \textbf{ ACC } & \textbf{ ACC } & \textbf{AUC} \\
\Xhline{1.5px}

InternVL3-78B~\cite{zhu2025internvl3} & \textbf{89.90} & \textbf{89.20} & \textbf{95.96} \\
InternVL3-1B~\cite{zhu2025internvl3} & 73.90 & 84.75 & 91.85 \\\hline
BLIP-Base~\cite{li2022blip} & 64.25 & 66.25 & 74.03 \\
BLIP-Large~\cite{li2022blip} & 59.10 & 65.60 & 72.48 \\\hline
BLIPv2-t5-xl~\cite{li2023blip2} & 60.15 & 62.05 & 65.47 \\
BLIPv2-t5-xl-coco~\cite{li2023blip2} & 63.15 & 64.70 & 70.10 \\
BLIPv2-t5-xxl~\cite{li2023blip2} & 61.85 & 62.45 & 67.99 \\
BLIPv2-opt-2.7b~\cite{li2023blip2} & 61.75 & 62.00 & 66.74 \\
BLIPv2-opt-6.7b~\cite{li2023blip2} & 60.30 & 61.70 & 65.56 \\\hline
DeepSeekVL-1.3B~\cite{lu2024deepseek_vl} & 82.70 & 81.40 & 89.31 \\
DeepSeekVL2-Small~\cite{wu2024deepseek_vl2} & 72.45 & 75.45 & 81.88 \\
DeepSeekVL2-Tiny~\cite{wu2024deepseek_vl2} & 72.05 & 73.30 & 81.27 \\\hline
Chameleon-7b~\cite{Chameleon2024Meta} & 50.15 & 51.15 & 52.50\\
Chameleon-30b~\cite{Chameleon2024Meta} & 50.00 & 53.15 & 55.55 \\\hline
DINO-ViT-B16~\cite{caron2021dino} & 70.55 & 69.30 & 74.92 \\
DINOv2-Base~\cite{oquab2023dinov2} & 70.90 & 69.90 & 77.91 \\
DINOv2-Giant~\cite{oquab2023dinov2} & 76.95 & 77.10 & 83.40 \\
DINOv2-Large~\cite{oquab2023dinov2} & 72.90 & 69.75 & 77.76 \\
DINOv2-Small~\cite{oquab2023dinov2} & 67.85 & 71.00 & 77.55 \\\hline
SAM-Base~\cite{kirillov2023segment_anything} & 55.45 & 54.90 & 58.42 \\
SAM-Huge~\cite{kirillov2023segment_anything} & 51.15 & 54.45 & 55.64 \\
SAM-Large~\cite{kirillov2023segment_anything} & 50.90 & 55.70 & 57.01 \\\hline
OpenCLIP-B-32~\cite{Cherti_2023_CVPR_OPENCLIP} & 68.85 & 69.20 & 76.22 \\
OpenCLIP-H-14~\cite{Cherti_2023_CVPR_OPENCLIP} & 77.25 & 78.25 & 86.52 \\
OpenCLIP-G-14~\cite{Cherti_2023_CVPR_OPENCLIP} & 76.70 & 75.20 & 83.16 \\\hline
ViT-B-16~\cite{dosovitskiy2020image_vit} & 61.55 & 64.10 & 70.05 \\
ViT-B-16-384~\cite{dosovitskiy2020image_vit} & 68.10 & 66.95 & 75.16 \\
ViT-B-32-384~\cite{dosovitskiy2020image_vit} & 65.10 & 64.85 & 71.61 \\
ViT-H-14-21k~\cite{dosovitskiy2020image_vit} & 66.95 & 67.10 & 74.15 \\
ViT-Hybrid-384~\cite{dosovitskiy2020image_vit} & 69.75 & 69.65 & 77.11 \\
ViT-L-16-224~\cite{dosovitskiy2020image_vit} & 67.05 & 64.90 & 70.94 \\
ViT-L-16-384~\cite{dosovitskiy2020image_vit} & 68.80 & 67.00 & 73.54 \\
ViT-L-32-384~\cite{dosovitskiy2020image_vit} & 63.70 & 64.75 & 70.21 \\\hline
ALIGN~\cite{jia2021scaling_align} & 68.20 & 69.05 & 77.45 \\\hline
LLaVA-1.5~\cite{llava15} & 68.10 & 74.65 & 82.77 \\
LLaVA-v1.6~\cite{liu2024llavanext} & 74.00 & 77.35 & 85.00 \\\hline
CLIP (B-16)~\cite{clip_radford2021learning} & 82.60 & 82.50 & 90.01 \\
CLIP (B-32)~\cite{clip_radford2021learning} & 73.80 & 79.00 & 87.46 \\
CLIP (L-14)~\cite{clip_radford2021learning} & 84.10 & 82.50 & 91.19 \\
CLIP (L-14-336)~\cite{clip_radford2021learning} & 85.25 & 84.80 & 92.45 \\
\Xhline{1.5px}
\end{tabular}}
\label{tab:deepfakes_results}
\end{table} 
Our goal is to measure how well general purpose multimodal large language models perform on the problem of DeepFake detection without any dedicated training, and whether these models are able to extract features that help differentiate fake images from the real ones.

To get a sense of how well these models understand what ``fake'' and ``real'' mean, we selected a set of state-of-the-art performing multimodal LLMs. We used the vision part of these models to encode the training dataset we sampled ten thousand images from the FaceForensics dataset~\cite{roessler2019faceforensicspp}. Then, used those extracted features to train a simple MLP classification head that consists of two linear layers, as well as a logistic regression classifier, and then measured their performance on the test set which consists of two thousand images. We trained the classification head for only 10 epochs with simple hyper-parameters; a learning rate of $0.001$, Adam optimizer with betas = $0.9$, $0.999$, and batch size of $32$. Results are reported in Table~\ref{tab:deepfakes_results}. The test set used for this measurement contains $2000$ images in which half are real and half are fake.

\begin{figure}[!ht]
    \centering
    \includegraphics[width=\linewidth]{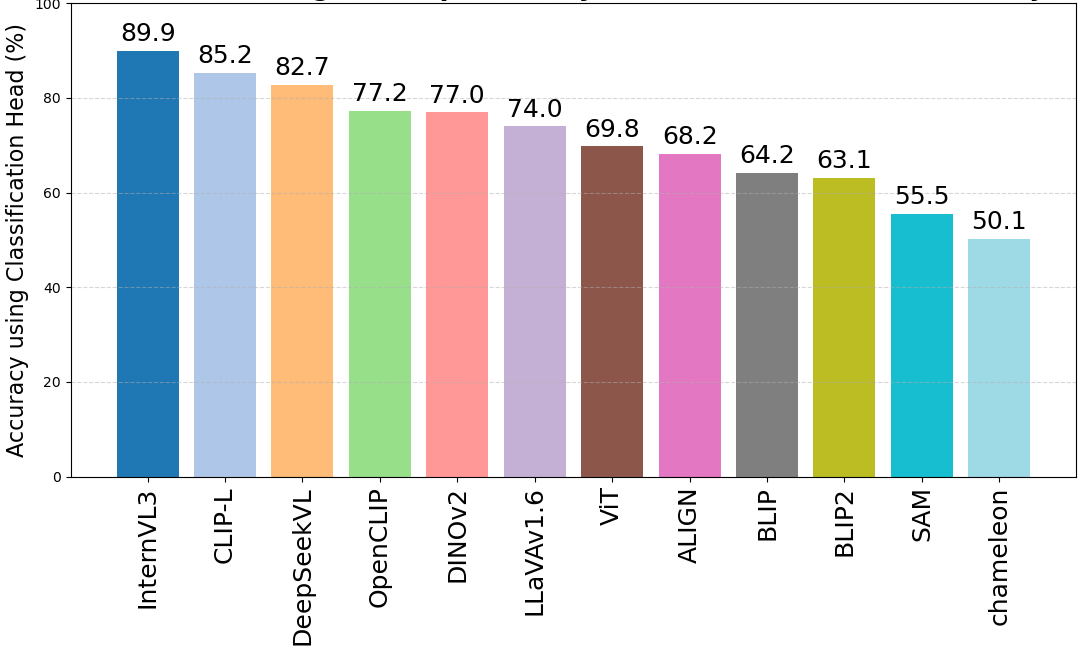}
    \caption{DeepFake detection accuracy by training and cross validating classification head with features extracted by Vision Language Models.
    }
    \label{fig:barplot_for_deepfakes}
\end{figure}

As illustrated, \textit{InternVL3-78B} achieves the highest performance, reaching nearly $90\%$ accuracy with a simple MLP classification head trained in just a few minutes. This result underscores the strength of the pretrained features extracted by the model and highlights the potential of leveraging such foundation MLLMs for DeepFake detection. The fact that such high accuracy is attainable with minimal training suggests that even greater performance could be achieved using more sophisticated classifiers or deeper neural networks built on top of these representations.

We present the comparative DeepFake detection accuracies in Figure~\ref{fig:barplot_for_deepfakes}. The superior performance of InternVL3-78B~\cite{zhu2025internvl3} can be attributed to two key factors: the incorporation of random JPEG compression during training and its unified end-to-end training strategy. Unlike models that pretrain individual components (e.g., vision encoders, language models, and multimodal projectors) in isolation, InternVL3-78B jointly optimizes all components—vision tower, multimodal projector, and LLM—on multimodal data. This integrated training approach enables the model to learn richer visual representations by leveraging cross-modal supervision, thereby enhancing its effectiveness in detecting subtle artifacts indicative of synthetic content such as DeepFakes.

\subsection{Morph Attack Detection}
\begin{table*}[!ht]
\centering
\small
\caption{EER (\%) / BPCER@10\%APCER (\%) for different models across six face morph datasets}
\resizebox{1.00\textwidth}{!}{  
\begin{tabular}{l|cccccc}
\Xhline{1.5px}
\multicolumn{7}{c}{\textbf{Morph Attack Detection (Face)}} \\
\Xhline{1.5px}
\textbf{Model} & \textbf{AMSL} \cite{ref64}  & \textbf{Facemorpher} \cite{ref68}& \textbf{OpenCV} \cite{ref97} & \textbf{StyleGAN} \cite{ref69} & \textbf{Webmorph} \cite{ref70} & \textbf{MorDiff} \cite{ref9} \\
\Xhline{1.5px}
InternVL3-1B~\cite{zhu2025internvl3} & 52.20/75.68 & 36.60/14.78 & 63.90/88.2 & 49.65/45.83 & 23.69/7.77 & 33.87/22.19 \\\hline

LLaVA-1.5-7B & 56.97/16.92 & 66.66/40.25 & 56.12/16.2 & 68.95/44.84 & 66.59/38.13 & 56.87/17.69 \\
LLaVA-v1.6-M-7B& 55.20/17.33 & 81.38/69.69 & 58.14/24.2 & 78.66/67.22 & 65.53/36.01 & 51.87/7.70 \\

\hline
BLIP-base~\cite{li2022blip} & 55.05/43.77 & 68.04/51.93 & 64.35/58.4 & 62.73/57.14 & 52.95/25.70 & 49.70/25.14 \\
BLIP-large~\cite{li2022blip} & 55.15/40.00 & 61.99/56.65 & 64.53/55.8 & 61.16/47.07 & 40.10/10.88 & 49.51/31.70 \\

BLIP2-t5-xxl~\cite{li2023blip2} & 56.07/48.78 & 55.44/64.35 & 58.64/64.8 & 47.46/47.40 & 41.91/41.24 & 52.90/53.32 \\
BLIP2-t5-xl~\cite{li2023blip2} & 48.19/33.01 & 47.72/37.02 & 55.71/53.0 & 47.04/33.69 & 40.41/25.37 & 49.01/34.6 \\
BLIP2-t5-xl-coco~\cite{li2023blip2} & 47.98/34.57 & 53.60/36.89 & 55.77/46.2 & 51.33/32.37 & 42.72/24.06 & 48.25/28.17\\
BLIP2-6.7b~\cite{li2023blip2} & 54.06/58.62 & 46.31/45.09 & 49.87/53.2 & 38.13/30.72 & 42.84/36.17 & 51.08/53.64 \\

\hline
DINO-vitb16~\cite{caron2021dino} & 54.08/13.10 & 73.30/50.56 & 69.83/46.6 & 73.82/51.61 & 53.30/10.56 & 64.60/33.17 \\
DiNOv2-Base~\cite{oquab2023dinov2}  & 51.68/54.85 & 61.95/69.44 & 46.55/42.6 & 64.22/68.04 & 53.03/55.56 & 56.51/58.56 \\
DiNOv2-Small~\cite{oquab2023dinov2} & 59.78/79.95 & 58.63/83.60 & 53.33/73.0 & 49.82/64.00 & 45.17/50.74 & 52.92/68.22 \\
DiNOv2-Large~\cite{oquab2023dinov2}  & 43.98/26.57 & 53.11/39.88 & 52.04/31.8 & 57.32/34.43 & 46.45/29.54 & 49.28/26.29 \\
DiNOv2-Giant~\cite{oquab2023dinov2} & 54.56/26.94 & 61.09/42.98 & 58.28/33.4 & 58.93/34.68 & 49.18/19.15 & 57.06/30.96 \\
\hline

OpenCLIP-H-14~\cite{clip_radford2021learning}  & 58.76/17.89 & 71.88/56.10 & 53.54/56.1 & 75.52/56.98 & 67.93/43.78 & 52.85/10.65 \\
OpenCLIP-L-32~\cite{clip_radford2021learning} & 51.47/52.30 & 54.86/52.30 & 62.01/66.6 & 59.31/49.75 & 52.77/37.02 & 45.52/33.8 \\
OpenCLIP-base-16~\cite{clip_radford2021learning} & 52.70/38.07 & 62.54/56.77 & 66.73/60.2 & 67.68/65.07 & 57.53/43.78 & 56.73/42.18 \\
CLIP-L-14-336~\cite{clip_radford2021learning} & 57.00/18.94 & 78.82/59.63 & 57.42/18.8 & 76.34/56.65 & 66.76/46.40 & 53.11/7.21 \\
\hline

SAM-base~\cite{kirillov2023segment_anything} & 56.45/73.29 & 63.33/83.11 & 63.16/77.8 & 51.06/61.52 & 36.78/34.94 & 53.00/70.35 \\
SAM-huge~\cite{kirillov2023segment_anything} & 52.53/66.44 & 55.88/65.22 & 53.40/67.2 & 38.80/41.95 & 25.60/4.66 & 56.79/70.02 \\
SAM-ViT-Large~\cite{kirillov2023segment_anything} & 58.85/42.25 & 69.39/76.4 & 43.82/26.2 & 36.72/19.97 & 59.47/60.52 & 59.91/57.1 \\

\hline
Chameleon-7B~\cite{Chameleon2024Meta} & 50.00 / 100.00 & 50.00 / 100.00 & 50.00 / 100.00 & 50.00 / 100.00 & 50.00 / 100.00 & 50.00 / 100.00 \\
Chameleon-30B~\cite{Chameleon2024Meta} & 50.00 / 100.00 & 50.00 / 100.00 & 50.00 / 100.00 & 50.00 / 100.00 & 50.00 / 100.00 & 50.00 / 100.00 \\
\Xhline{1.5px}
\end{tabular}}
\label{tab:eer_bpcer}
\end{table*}

\begin{figure*}[!ht]
    \centering
    \makebox[\textwidth][c]{ \includegraphics[width=1.05\textwidth]{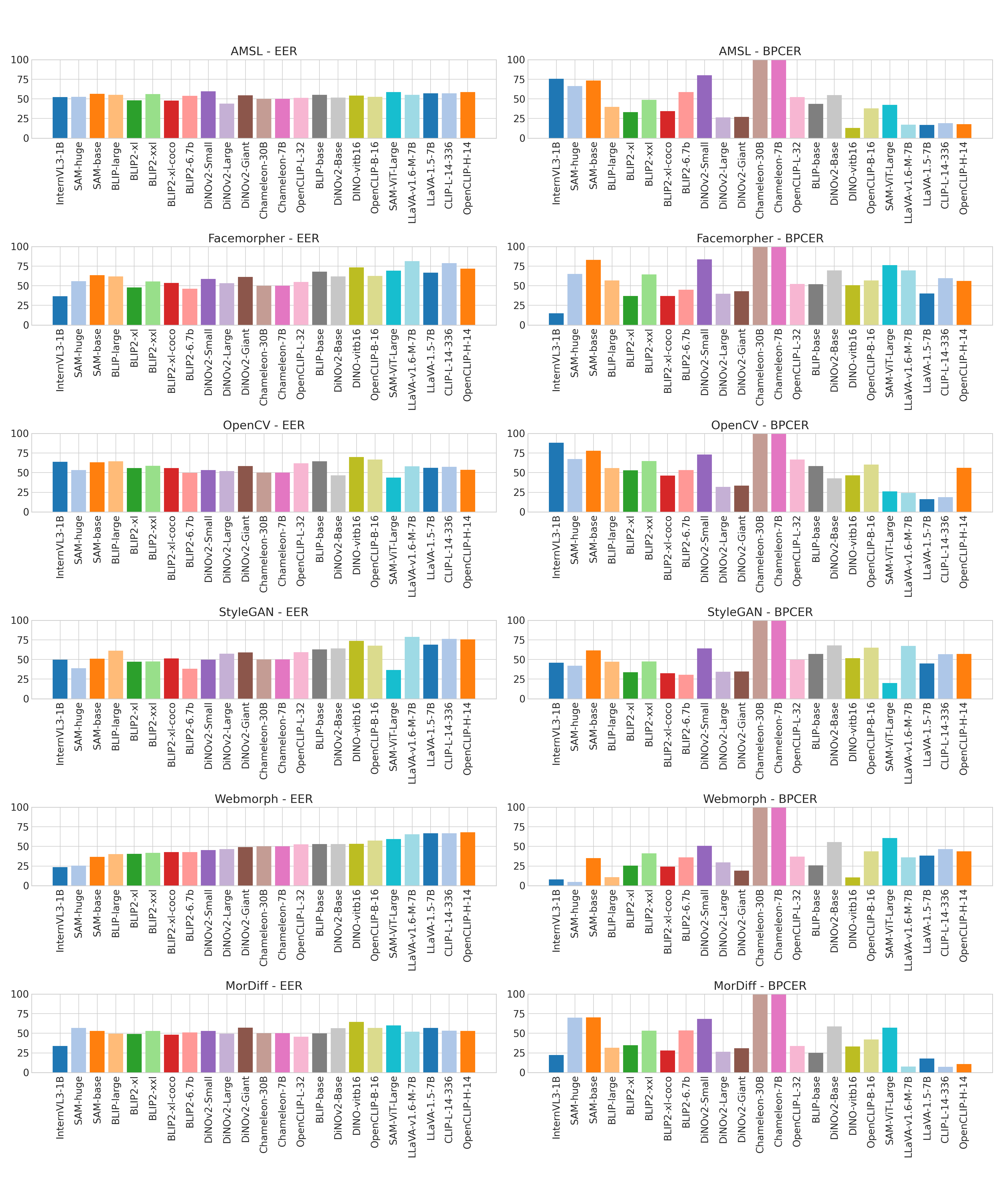}}
    \caption{Equal Error Rate (EER)(\%) and BPCER@10\%APCER for Morph Attack Detection for different Vision Language Models.  The lower these numbers, the better the detector.}
    \label{fig:morph_barplot}
\end{figure*}

Table~\ref{tab:eer_bpcer} presents a comprehensive comparison of Equal Error Rate (EER) and BPCER at $10\%$ APCER across 23 vision-language  models (VLM's) evaluated on six  face morphing datasets comprising landmarks and deep-learning based morphs. The results indicate considerable variability in detection performance across both models and datasets. InternVL3-1B~\cite{zhu2025internvl3} consistently achieves strong performance, particularly on the WebMorph~\cite{ref70} and faceMorpher~\cite{ref68} datasets, with EER/BPCER scores as low as 23.69\%/7.77\% and 36.60\%/14.78\%, respectively. Among the LLaVA variants, both perform poorly on both metrics with best BPCER of 7.70\% on Mordiff dataset. 
BLIP2 variants also perform poorly on all datasets with the best BPCER of 10.88\% in WebMorph dataset.
Surprisingly, the Chameleon models perform at chance level across all datasets, indicating a lack of sensitivity to morphing artifacts. Figure \ref{fig:morph_barplot} summarizes the EER and and BPCER@10\%APCER. These findings highlight the challenge of generalizable morph detection and we do not recommend using \textit{vanilla} models for Morph Attack Detection. The challenge lies in identifying the fine-grained subtle differences between the morphs and non-morphs.

\section{Findings}

Several foundation Vision-Language Models (VLMs) demonstrated remarkably strong performance on face recognition benchmarks, notably OpenCLIP~\cite{Cherti_2023_CVPR_OPENCLIP}, BLIP~\cite{li2022blip}, CLIP~\cite{clip_radford2021learning}, and LLaVA~\cite{llava15}, all exceeding $90\%$ TMR@1\%FMR. This level of accuracy is particularly impressive given that these models were not explicitly trained for face recognition tasks. A common factor among these high-performing models is their reliance on language-guided training rather than the traditional supervised metric learning approach that is normally used by state of the art face recognition models~\cite{Wang2018Cosface:Recognition, deng2019arcface, kim2022adaface, meng2021magface}. Specifically, CLIP was trained using self-/weakly supervised contrastive learning, BLIP employed a semi-supervised approach combining multiple pretraining objectives, and LLaVA was instruction-tuned using GPT-generated question-answer pairs. These findings underscore the powerful representational capabilities learned through self- and semi-supervised paradigms~\cite{balestriero2024the}, which enable effective generalization to downstream biometric tasks without task-specific fine-tuning. In addition to them, those models perform very well when a lightweight classifier is trained with extracted features for soft-biometric attribute classification. 


Several Vision–Language Models (VLMs) demonstrate good generalization capabilities for iris recognition and iris presentation attack detection (PAD) tasks, despite being pretrained on unrelated objectives. As shown in Table~\ref{tab:iris-recognition-vlm}, DINO-ViTB16, achieves the highest TMR@1\%FMR across all four benchmarks (UND-Full, UND-Cropped, IITD-R-Full, IITD-R-Cropped), exceeding 96\% on the IITD-R subsets. These models employ self-supervised training, which likely contributes to their robust visual representations and superior performance in iris recognition. DeepSeek models show high recognition capability, particularly on the IITD-R benchmarks, suggesting that its large-scale pretraining is well aligned with biometric patterns.

Table~\ref{tab:iitd-pa-table} presents PAD performance on the IITD-P PA dataset using both zero-shot similarity classification (Acc) and trained classifiers (SVM, RBF, and Logistic Regression). Here, the DINOv2 models again stand out, with DINOv2-Base and DINO-ViT-B16 achieving near-perfect accuracy $(>99\%)$ after classifier training. In contrast, the BLIP2-flan-t5 variants perform modestly in zero-shot settings, but excel when paired with a linear SVM, reaching the highest SVM accuracy $(99.57\%)$ among the models. These results suggest that VLMs can serve as effective feature extractors for biometric tasks, and their representations—especially when coupled with lightweight classifiers—yield competitive PAD performance without task-specific fine-tuning. 

For DeepFake detection, the InternVL3-78B model achieved the highest accuracy using a simple MLP head without any architectural modifications, highlighting the strong representational capabilities of multimodal large language models (MLLMs) in this domain. Similarly, in the task of morph attack detection, InternVL3-1B consistently demonstrated superior performance, achieving notably low EER/BPCER scores on challenging datasets such as WebMorph and FaceMorpher. In contrast, models like LLaVA and BLIP2 showed limited effectiveness, and Chameleon models performed at chance level, underscoring both the complexity of detecting subtle morphing artifacts and the current limitations of vanilla vision-language models (VLMs) in this task.

\FloatBarrier
\section{Summary and Future Work}



This work offers an extensive evaluation of foundation Vision-Language Models (VLMs) across a diverse suite of biometric tasks, including face and iris recognition, soft biometric attribute classification (gender and ethnicity), and multiple forms of adversarial attack detection—namely, face morphing, DeepFakes. Our results demonstrate the strong zero-shot and few-shot capabilities of several VLMs, revealing their remarkable ability to extract discriminative, semantically rich feature embeddings from biometric data without any task-specific fine-tuning.

Notably, CLIP, OpenCLIP, and BLIPv2 exhibited consistently high accuracy in face recognition, while DINO and DINOv2 emerged as top performers in iris recognition and iris presentation attack detection, respectively. For deepfake and morph attack detection, the InternVL3 model achieved the best scores among the VLMs, underscoring its robustness across diverse biometric threat vectors.

These findings point to a possibility of a paradigm shift in biometric system design—from traditional supervised pipelines to language-supervised, self-supervised learning frameworks. Such models can alleviate the reliance on large-scale annotated datasets, which are often difficult and expensive to obtain in biometric domains. By leveraging the generalization power of VLMs trained on vast multimodal corpora, future biometric systems can benefit from more scalable, adaptable, and generalizable architectures, ultimately pushing the boundaries of what is achievable with limited supervision.

However, in the context of this paper, there is still more work to do. 
\begin{enumerate}
    \item Performing this analysis on other biometric modalities such as fingerprints, and on large-scale datasets with extreme diversity.
    \item Exploiting the text encoders in foundation models to obtain full multimodal synergy. 
    \item Enhancing the explainability and reasoning capabilities of these models specifically for biometrics. 
    \item Investigating the reason for disparate performance of different foundation models on each biometric task. 
    \item Understanding the content and nature of the training data used by these foundation models to assess bias and fairness; ensure accountability and transparency; evaluate data quality and relevance; identify legal and ethical issues, if any; and support model improvement and fine-tuning. 
    \item Determining effective ways to perform unlearning in order to purge incorrect concepts learned by these foundation models; improve model efficiency; and uphold privacy rights. 
    \item Applying knowledge distillation to foundation models thereby transferring knowledge from a large, often over-parameterized ``teacher" model to a smaller, more efficient ``student" model. 
    \item Fusing foundation models to further enhance the accuracy of biometric tasks and leverage their generalization capabilities by combining the strengths of multiple large-scale models to improve performance, robustness, and adaptability. 
\end{enumerate}

\section*{Ethical Considerations}
All experiments in this study were carried out on local offline servers with institutional infrastructure. Sensitive biometric data was not uploaded to commercial cloud services or external APIs thereby ensuring strict compliance with privacy and data protection standards.

\section*{Acknowledgments}
This work was supported in part through computational resources and services provided by the Institute for Cyber-Enabled Research (ICER) at Michigan State University (HPCC). Additional experiments were conducted on secure internal compute servers of the iPRoBe lab.

\balance
\bibliographystyle{IEEEtran}
\bibliography{bibliography}

\newpage

\end{document}